%% file: main-v5.tex
\documentclass{article}

\usepackage{PRIMEarxiv}

\usepackage[utf8]{inputenc} 
\usepackage[T1]{fontenc}    
\usepackage{amsmath}        
\usepackage{mathtools}      
\usepackage{amsthm}         
\usepackage{hyperref}       
\usepackage{url}            
\usepackage{booktabs}       
\usepackage{array}          
\usepackage{textcomp}      
\usepackage{amsfonts}       
\usepackage{amssymb}        
\usepackage{nicefrac}       
\usepackage{microtype}      
\usepackage{graphicx}
\usepackage[numbers]{natbib}
\usepackage{cleveref}
\usepackage{color}
\usepackage{multirow}
\usepackage{xspace}
\usepackage[normalem]{ulem} 
\usepackage{comment-macros}  
\graphicspath{{media/}}     

\usepackage{listings}       
\input{symbols.tex}              
\input{sections-v5/symbols-v5.tex} 

\makeatletter
\newcommand{\dl}[1]{{\color{red}%
  \let\dl@citet\citet\let\dl@cite\cite\let\dl@Cref\Cref\let\dl@cref\cref\let\dl@ref\ref%
  \renewcommand{\citet}[2][]{\mbox{\dl@citet{##2}}}%
  \renewcommand{\cite}[2][]{\mbox{\dl@cite{##2}}}%
  \renewcommand{\Cref}[1]{\mbox{\dl@Cref{##1}}}%
  \renewcommand{\cref}[1]{\mbox{\dl@cref{##1}}}%
  \renewcommand{\ref}[1]{\mbox{\dl@ref{##1}}}%
  \sout{#1}}}
\makeatother

\makeatletter
\newcommand{\cldl}[1]{{\color{blue}%
  \let\cldl@citet\citet\let\cldl@cite\cite\let\cldl@Cref\Cref\let\cldl@cref\cref\let\cldl@ref\ref%
  \renewcommand{\citet}[2][]{\mbox{\cldl@citet{##2}}}%
  \renewcommand{\cite}[2][]{\mbox{\cldl@cite{##2}}}%
  \renewcommand{\Cref}[1]{\mbox{\cldl@Cref{##1}}}%
  \renewcommand{\cref}[1]{\mbox{\cldl@cref{##1}}}%
  \renewcommand{\ref}[1]{\mbox{\cldl@ref{##1}}}%
  \sout{#1}}}
\makeatother

\definecolor{rvorange}{rgb}{0.90,0.42,0.00}
\definecolor{rvpurple}{rgb}{0.50,0.00,0.63}

\makeatletter
\newcommand{\rvdl}[1]{{\color{rvorange}%
  \let\rvdl@citet\citet\let\rvdl@cite\cite\let\rvdl@Cref\Cref\let\rvdl@cref\cref\let\rvdl@ref\ref%
  \renewcommand{\citet}[2][]{\mbox{\rvdl@citet{##2}}}%
  \renewcommand{\cite}[2][]{\mbox{\rvdl@cite{##2}}}%
  \renewcommand{\Cref}[1]{\mbox{\rvdl@Cref{##1}}}%
  \renewcommand{\cref}[1]{\mbox{\rvdl@cref{##1}}}%
  \renewcommand{\ref}[1]{\mbox{\rvdl@ref{##1}}}%
  \sout{#1}}}
\makeatother

\definecolor{edburgundy}{rgb}{0.55,0.00,0.16}

\makeatletter
\newcommand{\eddl}[1]{{\color{edburgundy}%
  \let\eddl@citet\citet\let\eddl@cite\cite\let\eddl@Cref\Cref\let\eddl@cref\cref\let\eddl@ref\ref%
  \renewcommand{\citet}[2][]{\mbox{\eddl@citet{##2}}}%
  \renewcommand{\cite}[2][]{\mbox{\eddl@cite{##2}}}%
  \renewcommand{\Cref}[1]{\mbox{\eddl@Cref{##1}}}%
  \renewcommand{\cref}[1]{\mbox{\eddl@cref{##1}}}%
  \renewcommand{\ref}[1]{\mbox{\eddl@ref{##1}}}%
  \sout{#1}}}
\makeatother

\definecolor{grgreen}{rgb}{0.00,0.45,0.20}

\makeatletter
\newcommand{\grdl}[1]{{\color{grgreen}%
  \let\grdl@citet\citet\let\grdl@cite\cite\let\grdl@Cref\Cref\let\grdl@cref\cref\let\grdl@ref\ref%
  \renewcommand{\citet}[2][]{\mbox{\grdl@citet{##2}}}%
  \renewcommand{\cite}[2][]{\mbox{\grdl@cite{##2}}}%
  \renewcommand{\Cref}[1]{\mbox{\grdl@Cref{##1}}}%
  \renewcommand{\cref}[1]{\mbox{\grdl@cref{##1}}}%
  \renewcommand{\ref}[1]{\mbox{\grdl@ref{##1}}}%
  \sout{#1}}}
\makeatother

\definecolor{smteal}{rgb}{0.00,0.50,0.55}
\newcommand{\sm}[1]{\textcolor{smteal}{#1}}

\theoremstyle{definition}
\newtheorem{definition}{Definition}
\newtheorem{example}{Example}
\theoremstyle{plain}

\crefname{lstlisting}{Listing}{Listings}
\Crefname{lstlisting}{Listing}{Listings}

\title{Polyscene: Scenario Planning as Diversity Planning}

\author{
  Mustafa F. Abdelwahed, Joan Espasa \\
  School of Computer Science \\
  University of St Andrews \\
  St Andrews\\
  \texttt{\{ma342, jea20\}@st-andrews.ac.uk} \\
   \And
  Felipe Meneguzzi \\
  School of Natural and Computing Sciences \\
  University of Aberdeen \\
  Aberdeen\\
  \texttt{felipe.meneguzzi@abdn.ac.uk} \\
}

\begin{document}
\maketitle

\begin{abstract}
\input{sections-v5/00-abstract.tex}
\end{abstract}

\keywords{Automated Planning \and Diversity Planning \and Scenario Planning}

\input{sections-v5/01-introduction.tex}
\input{sections-v5/02-background.tex}
\input{sections-v5/03-engine.tex}
\input{sections-v5/04-case-studies.tex}
\input{sections-v5/04b-related-work.tex} 
\input{sections-v5/05-conclusion.tex}

\bibliographystyle{unsrtnat}
\bibliography{references}

\end{document}

%% file: symbols.tex
\newcommand{\fluentdomainname}{\operatorname{Range}}
\newcommand{\fluentdomain}[1]{\fluentdomainname(#1)}  

\newcommand{\allscenarios}{\Pi^{\mathit{sc}}_{\planningtask}} 

\newcommand{\cartelname}{OPEC\xspace}
\newcommand{\importername}{the United States\xspace}
\newcommand{\Importername}{The United States\xspace}
\newcommand{\clientname}{Israel\xspace}
\newcommand{\rivalname}{Egypt\xspace}

\newcommand{\fbixoperator}[1]{\operatorname{FBI}_\mathit{#1}}

\newcommand{\planningtask}{\Xi}
\newcommand{\fluent}{v}
\newcommand{\allfluents}{V}
\newcommand{\allstate}{S}
\newcommand{\state}[1]{s_{#1}}
\newcommand{\allaction}{A}
\newcommand{\action}[1]{a_{#1}}
\newcommand{\transitionfn}{\gamma}
\newcommand{\costfn}{\operatorname{cost}}
\newcommand{\initialstate}{I}
\newcommand{\goalstate}{G}
\newcommand{\singleplan}[1]{\pi_{\mathit{#1}}}
\newcommand{\allplans}{\Pi_\planningtask}
\newcommand{\diversesetplan}[1]{\Psi_{#1}}
\newcommand{\planscount}{k}
\newcommand{\costthreshold}{c}
\newcommand{\distancethreshold}{d}

\newcommand{\plannername}[3]{\operatorname{#1}_{#2}^{#3}}

\newcommand{\fiplanner}[1]{\plannername{FI}{#1}{}}

\newcommand{\actionpre}{\operatorname{pre}}
\newcommand{\actioneff}{\operatorname{eff}}

\newcommand{\diversityindicator}{\lambda}

%% file: sections-v5/symbols-v5.tex
\newcommand{\concernset}{C}                               
\newcommand{\driverset}{D}                                
\newcommand{\outcome}[2]{\operatorname{out}_{#1}(#2)}     
\newcommand{\scenarioset}{\Psi}                           

\newcommand{\ltlnext}{\mathord{\circ}}                     
\newcommand{\ltluntil}{\mathbin{\mathcal{U}}}             
\newcommand{\ltlprev}{\mathord{\ominus}}                  
\newcommand{\ltleventually}{\Diamond}                     
\newcommand{\ltlalways}{\Box}                             
\newcommand{\satformula}[1]{\mathcal{E}_{#1}}              
\newcommand{\satinit}{\mathcal{I}}                         
\newcommand{\sattrans}{\mathcal{T}}                        
\newcommand{\satgoal}{\mathcal{G}}                         
\newcommand{\trajconstraints}{\Phi}                        
\newcommand{\expectedoutcome}{\widehat{\driverset}}  
\newcommand{\critformula}[1]{\varphi_{\mathrm{#1}}}      

\newcommand{\actorof}[1]{\mathrm{act}(#1)}                 
\newcommand{\targetof}[1]{\mathrm{tgt}(#1)}                
\newcommand{\actors}{\Gamma}                             
\newcommand{\objectives}{\mathcal{O}}                       
\newcommand{\objectiveof}[1]{u_{#1}}                        
\newcommand{\anactor}{g}                                   
\newcommand{\avalue}{x}                                    
\newcommand{\aweight}{w}                                   
\newcommand{\happenings}{H}                                
\newcommand{\allstates}{S}                                 
\newcommand{\classof}[1]{\mathrm{schema}(#1)}             
\newcommand{\sigof}[1]{\mathrm{sig}(#1)}                   
\newcommand{\ltlsince}{\mathbin{\mathcal{S}}}              
\newcommand{\ltlonce}{\mathord{\blacklozenge}}             
\renewcommand{\fluentdomainname}{\operatorname{range}}

\newcommand{\topic}{\mathbb{T}}                            

%% file: sections-v5/00-abstract.tex
Scenario planning is a foresight practice in which analysts construct a small set of plausible futures and prepare an organisation for each.
Each scenario is a narrative that must be plausible, internally consistent, relevant and challenging, and the set as a whole must be differentiated.
The practice remains manual:
a workshop (i.e., a facilitated meeting in which a small group of experts selects the drivers and writes the narratives) considers only the factors it has time for, carries the biases of the experts who run it, and cannot be redone whenever conditions change.
In this paper, we automate scenario planning by casting it as a diversity planning problem, in which two scenarios differ when they end in different narratives on the analyst's drivers.
The result is a set of scenarios, one for each narrative that the topic can reach within a budget (i.e., a bound on a narrative's cost).
Our implementation, Polyscene, is a generic engine that generates scenarios from a topic (i.e., a planning domain model of the subject. We use geopolitical conflicts and shipping-route disruptions as case studies.
For the first, we build the planning problem's initial state from the GDELT Project so it reflects current relationships among the countries involved.
Polyscene then generates scenarios showing how relationships among those countries change. With a board of six countries as of 1 January 2021 and a budget of five events, the conflict set puts five of the 15 country pairs at war, and each of the three wars that break out among these countries by 1 January 2026 ends a scenario.
For the second, we build the initial state from ship traffic, a supply-pressure index, and the news, and Polyscene generates scenarios showing what a closed or diverted route could do to a manufacturer's parts stock. With boards of four routes as of the same date and the same budget, each set holds three outcomes (e.g., the route closed, the lines diverted, or both), and the Suez Canal set holds the blockage of March 2021 and the diversion of container lines from the Red Sea from December 2023.

%% file: sections-v5/01-introduction.tex
\section{Introduction}\label{sec:introduction}

Scenario planning is a foresight method in which analysts prepare for the future without trying to forecast it~\cite{wack1985scenarios,schoemaker1995scenario}.
They develop a limited set of credible alternative futures, called scenarios, and the organisation prepares to respond to each one.
Each scenario is a narrative explaining how current conditions might unfold into a particular future.
The approach gained prominence at Shell in the early 1970s.
Its planning group set a family of scenarios ranging from producer states restricting oil supply to one in which no crisis occurred and growth continued~\cite{wack1985scenarios}.
When the 1973 embargo occurred, Shell's leadership had already worked through the disruption storyline~\cite{wack1985scenarios,vanderheijden1996scenarios}.
Today, analysts typically convene in workshops (i.e., facilitated meetings in which a small group of experts selects the drivers and writes the narratives) to identify key drivers of change, craft a small set of narratives, and assess them against the discipline's criteria~\cite{amer2013review,types-scenario-planning}.
This approach has three shortcomings that argue for automation.
First, a workshop considers only a limited number of factors and produces three to five narratives~\cite{amer2013review}, so some futures are never explored.
Second, it is vulnerable to expert bias.
Scenario planning is meant to counter biases such as tunnel vision and anchoring, and the evidence that it does so is mixed~\cite{types-scenario-planning}, so a workshop can still fall prey to them.
Tunnel vision directs attention towards futures that feel most familiar, while anchoring pulls each narrative towards the initial assumption encountered.
Third, the exercise is infeasible to redo each time conditions change~\cite{davis2023towards,thesis-scenario-acceleration}.
These shortcomings leave structured scenario planning largely limited to organisations that can support dedicated foresight teams, while smaller organisations and public-sector bodies conduct it infrequently~\cite{Kodding2023}.
Our approach addresses these limitations in three ways.
First, it can account for more factors than a workshop. A workshop usually builds its stories around a small set of drivers, often focusing on the two considered most uncertain, then combining them to produce four scenario themes~\cite{schoemaker1995scenario}. In one conflict topic case study, it aims to predict future geopolitical conflicts. 
The topic models conflict as pairs between countries, so the space of possible narratives is $2^{15}$ war combinations rather than just four themes. The engine then explores all narratives allowed by the topic's events within the budget and produces one scenario for every combination it can reach. 
Second, our approach does not rank narratives by cost, likelihood, or other soft factors, so familiar futures have no search advantage over unfamiliar ones. Third, the analyst can rerun it whenever conditions change, and the same topic then returns a new set.
Previous work has automated certain aspects of scenario planning.
What remains missing is a trajectory-level criterion that defines when a generated chain of events counts as a scenario.
For example, the IBM Scenario Planning Advisor (SPA) scores each generated plan using a cost derived from the estimated likelihoods, then clusters these plans into scenarios~\cite{sohrabi2018aaai,sohrabi2018ibm,feblowitz2021ibm,katz2018topk}.
Another line of work, such as cross-impact balance and morphological analysis, finds driver combinations that satisfy a consistency metric~\cite{weimerjehle2006cross,ritchey2006problem}.
Although these methods state what makes a driver combination acceptable, they return an end state rather than a sequence of events.
More recent studies use large language models to draft foresight scenarios~\cite{fi17010048} and, in other fields, to generate test scenarios for autonomous driving~\cite{llmscenario}. While these narratives read smoothly, they often fail to remain consistent~\cite{simon2024large}, and hallucinations weaken their credibility~\cite{bang2025hallulens}. 

Our approach frames scenario planning as diversity planning, with one change.
Instead of specifying a single goal state, we impose a budget on the plan's cost.
Any state reachable within that limit becomes a possible future, and any plan counts as a narrative.
Not every narrative is a scenario (i.e., plausible, internally consistent, relevant, and challenging), so we encode the scenario criteria into the planning problem (\Cref{sec:engine}).
The workflow starts with developing the planning domain (i.e., the fluents and actions of the subject). The organisation's question then turns the domain into a topic (i.e., the domain together with the actors' objectives, constraints, concerns, and drivers that the criteria read).
A script grounds the topic's initial state from the record, and our engine then returns one scenario for each narrative that the topic can reach within the budget.
We implement our approach, Polyscene, in Answer Set Programming~\cite{gebser2012answer}\footnote{Code will be publicly available upon publication.}.
The engine is generic because the criteria are stated over the structure that every planning problem has (i.e., who acts on whom, what an action reads and changes, and what the actors want).
This means that they can be implemented once and the topic supplies the planning domain.
In summary, the main contributions of this paper are: (1) a formulation of scenario planning as a diversity planning problem, (2) a generic engine, called Polyscene, that generates one scenario per narrative a given topic can reach within the budget (\Cref{sec:engine:sat,sec:engine:differentiation}), and (3) two case studies, on geopolitical conflicts and on supply chains (\Cref{sec:case-study}). As of 1 January 2021, the conflict set includes 8 narratives within a budget of 5 events, and each of the three wars that broke out among these countries by 1 January 2026 ends a scenario. On four shipping-route boards as of the same date, each supply-chain set includes three outcomes (e.g., the route closed, the lines diverted, or both). The Suez Canal set holds the blockage of the canal in March 2021 and the diversion of container lines from the Red Sea from December 2023, and it misses the production stops of January 2024. \Cref{sec:background} reviews scenario planning and automated planning, using Shell's exercise as a running example; \Cref{sec:related-work} compares our approach with prior work on automating scenario planning; and \Cref{sec:conclusion} discusses the findings and suggests future work.

%% file: sections-v5/02-background.tex
\section{Background}
\label{sec:background}

 \Cref{sec:background:scenario-planning} explains how analysts practise scenario planning and outlines the criteria they use to evaluate a scenario, using Shell's 1972 exercise as a running example.
 \Cref{sec:background:planning} defines the planning problem, whose plans are our scenario narratives, and frames Shell's exercise as an instance of it.
 \Cref{sec:background:planners} defines the diversity planning problem and reviews the diverse planners that solve it, which form the basis for the engine described in \Cref{sec:engine}.

\subsection{Scenario Planning}\label{sec:background:scenario-planning}

A scenario planning exercise begins by identifying the factors that influence a given topic's future.
A subset of these factors is then selected as \emph{scenario drivers} (i.e., the factors whose values distinguish one future from another)~\cite{types-scenario-planning}.
Each scenario is then developed into a narrative describing how the present situation could evolve.
\citet{kahnwiener1967year} define a scenario as a hypothetical sequence of events constructed to focus attention on causal processes and decision points.
Analysts evaluate scenarios by several criteria, including plausibility, internal consistency, relevance and challenge~\cite{amer2013review,vanderheijden1996scenarios}.
A fifth criterion applies to the set rather than a single scenario: the scenarios must be differentiated~\cite{amer2013review}.

 A scenario is \emph{plausible} if it can happen~\cite{wilson1998mental}.
 This does not imply it must be likely~\cite{kosow2008methods,ramirez2014plausibility}.
 Plausibility does not rank scenarios, so we view all scenarios within a set as equally plausible~\cite{amer2013review,types-scenario-planning}.
 A scenario is \emph{internally consistent} when the logics it combines are consistent~\cite{wilson1998mental}.
 A scenario is \emph{relevant} when it offers specific insights into the future that bear on the decision at hand~\cite{wilson1998mental}.
 This means that relevance depends on the audience, since it is judged against the concerns of the organisation for which the scenario is written~\cite{vanderheijden1996scenarios,schoemaker1995scenario}.
 A scenario is considered \emph{challenging} if it challenges the organisation's conventional expectation about the future~\cite{wilson1998mental}, which \citet{vanderheijden1996scenarios} describe as presenting a new viewpoint on the matters at hand.
 Finally, a collection of scenarios is \emph{differentiated} when its scenarios differ structurally rather than as variations on one base case~\cite{wilson1998mental}.
 To clarify these requirements, we use Shell's 1972 exercise as our running example throughout this paper. Note that \Cref{ex:shell-setting}'s four outcomes represent final configurations, each compatible with numerous distinct trajectories.
An analyst cannot write out all these trajectories by hand, motivating automation.
\Cref{sec:related-work} reviews earlier efforts to automate parts of this exercise.

\begin{example}[Shell's exercise]\label{ex:shell-setting}
We use a simplified version of Shell's exercise, set in 1972 before the October 1973 Egyptian--Israeli war broke out.
The actors and events below are ours and follow the 1973 embargo; Shell's own scenarios focused on producer revenues and oil supply rather than war~\cite{wack1985scenarios}.
The key participants include the producer cartel (\cartelname), an oil-importing country that supports a regional ally (\importername), the ally itself (\clientname), and a neighbouring state that is hostile to the ally (\rivalname).
\Importername is allied with \clientname, while \clientname and \rivalname are adversaries.
The cartel operates as a unified, coordinated entity.
Each participant has a certain capacity to endure economic harm.
For instance, the cartel relies on oil revenues, while the importer depends on its ability to finance oil purchases.
Two elements of this situation are identified as scenario drivers: (i) whether the cartel maintains coordinated behaviour, and (ii) whether the region remains peaceful or descends into war.
These drivers produce four possible outcomes, and we examine two, as Shell's team examined two families of futures.
In the disruption narrative, war breaks out, the importing country resupplies its ally, and the still-coordinated cartel responds with an embargo that severely restricts the importer's capacity.
In the continuity narrative, coordination among cartel members deteriorates, the region remains peaceful, and economic growth continues.
We leave the remaining two outcomes unexplored, as an analyst would consider them less insightful.
One of these unexamined outcomes features a united cartel in a peaceful region, which Shell's surprise-free scenario (i.e., the future in which nothing disturbs the present trends) took as the projection at the time~\cite{wack1985scenarios}.
\end{example}

\subsection[Narrative Planning]{Narrative Planning}\label{sec:background:planning}

\Cref{sec:background:scenario-planning} outlined scenario planning and the criteria any scenario must fulfill.
This subsection defines the planning problem the engine builds on.
Automated planning studies how to find a sequence of actions that takes the world from an initial state to one that satisfies a goal~\cite{ghallab2016automated}.
Our setting differs from this in two ways.
We set the goal to $\goalstate=\top$ (i.e., every final state satisfies it) and instead bound a narrative's cost, because a scenario exercise has no single goal state, as it explores a potential future.
Our setting also needs to rule out entire sequences of actions, which preconditions cannot do, because a precondition is checked only in the state where its action occurs.
In Shell's exercise, the cartel may split after it imposes an embargo, since it is coordinated at both moments, yet the embargo stays in force once the cartel splits, and an embargo enforced by a divided cartel is contradictory.
Adding a negated embargo to the precondition of the split would have to be repeated for every action that could divide the cartel, whereas a trajectory constraint (i.e., a formula over the whole sequence of states and actions) states the invariant once (i.e., an embargo in force requires a coordinated cartel) and judges every state the sequence passes through.
Admitting actions as atoms extends beyond PDDL3, which focuses on state constraints, and aligns with LTL action theories, where an action is a proposition related to its resulting situation\sm{~\cite{calvanese2002reasoning}}. 
This approach combines PDDL3's state constraints with the action-trajectory constraints \citet{bonassi2022action} added to PDDL3, which range over actions alone.
Thus, we model these invariants and scenario criteria as constraints and preferences in PDDL3~\cite{gerevini2009deterministic}, forcing the planner to consider the states it passes through to satisfy them. Following \citet{wright2018compiling,degiacomo2013ltlf}, we express the planning problem and its solution as follows.

\begin{definition}[Planning problem]\label{def:planning-problem}
 A \emph{planning problem} is a tuple
 $\planningtask=\langle\allfluents,\fluentdomainname,\allaction,\transitionfn,\costfn,\initialstate,\goalstate,\trajconstraints\rangle$, where:
 \begin{itemize}
 \item $\allfluents$ is a finite set of fluents (state variables).

 \item $\fluentdomainname$ is a family of finite sets of \emph{values}, one for each
 fluent; $\fluentdomain{\fluent}$ is the set of values for fluent $\fluent$. A
 \emph{state} is a valuation assigning each $\fluent\in\allfluents$ a value
 $\state{}[\fluent]\in\fluentdomain{\fluent}$, and the \emph{state space}
 $\allstate$ is the set of all of them.

 \item $\allaction$ is a finite set of \emph{actions}. Each action
 $\action{}=\langle\actionpre_{\action{}},\actioneff_{\action{}}\rangle$ has a
 \emph{precondition} $\actionpre_{\action{}}$, a logical formula\footnote{We allow
 $\land$, $\lor$, $\lnot$, $\implies$, and (in)equalities between values (if
 $\fluentdomain{\fluent}$ is totally ordered).} over $\allfluents$, and an \emph{effect}
 $\actioneff_{\action{}}$, a valuation assigning a value in $\fluentdomain{\fluent}$
 to each fluent $\fluent$ of a subset $\allfluents_{\action{}}\subseteq\allfluents$.
 Each action either has an \emph{actor} $\actorof{\action{}}$, the agent that performs it, or is a \emph{happening} (i.e., an action that no agent performs). In both cases, an action may be directed at a \emph{target} $\targetof{\action{}}$.
 Actions that differ only in the people and things they involve form a \emph{schema}. The symbol $\classof{\action{}}$ represents the schema of the action $\action{}$. We call $\sigof{\action{}}=\langle\classof{\action{}},\actorof{\action{}},\targetof{\action{}}\rangle$ the \emph{signature} of an action (i.e., its schema, actor and target, with the other parameters left out)

 \item $\transitionfn$ is a partial transition function $\transitionfn:\allstate\times\allaction\rightharpoonup\allstate$. It is defined for a state $\state{}$ and an action $\action{}$ if and only if $\state{}\models\actionpre_{\action{}}$ (i.e., $\action{}$ is applicable in $\state{}$), and it returns the state $\transitionfn(\state{},\action{})[\fluent]=\begin{cases}
 \actioneff_{\action{}}[\fluent] & \text{if } \fluent\in\allfluents_{\action{}},\\
 \state{}[\fluent] & \text{otherwise.}
 \end{cases}$

 \item $\costfn:\allaction\rightarrow\mathbb{R}_{\geq 0}$ is a \emph{cost function} assigning a non-negative cost to each action.

 \item $\initialstate\in\allstate$ is the \emph{initial state}.

 \item $\goalstate$ is the \emph{goal}, a logical formula over $\allfluents$.

 \item $\trajconstraints$ is a finite set of \emph{trajectory constraints} (i.e., LTL$_f$ formulas over the states that a plan passes through and the actions it takes). Each constraint $\theta$ is generated by the grammar $\theta::=\varphi\mid\neg\theta\mid\theta\wedge\theta\mid\ltlnext\theta\mid\theta\ltluntil\theta\mid\ltlprev\theta\mid\theta\ltlsince\theta$, starting from logical formulas $\varphi$ over $\allfluents\cup\allaction$ (i.e., formulas over the fluents in which an action $\action{}\in\allaction$ can also occur as an atomic proposition). A constraint is evaluated over the \emph{positions} of a sequence of actions $\langle\action{1},\ldots,\action{m}\rangle$ applied from $\initialstate$: position $0$ holds the initial state alone, and position $i\geq1$ holds the state reached after $\action{i}$ together with $\action{i}$ itself~\cite{wright2018compiling}.
 A formula $\varphi$ holds at a position when the state there satisfies it, and an action atom $\action{}$ is true exactly at the position of that action. \Cref{tab:ltl-operators} gives the meaning of each temporal operator at a position $i$.
 A sequence satisfies a constraint, written $\singleplan{}\models\theta$, when $\theta$ holds at position $0$.

 \end{itemize}
\end{definition}

\begin{table}
\centering
\footnotesize
\renewcommand{\arraystretch}{1.25}
\begin{tabular}{@{}llp{9.2cm}@{}}
\toprule
Operator & Name & Holds at position $i$ when \\
\midrule
$\ltlnext\theta$ & next & $\theta$ holds at $i+1$; false at the last position \\
$\theta_1\ltluntil\theta_2$ & until~\cite{degiacomo2013ltlf} & $\theta_2$ holds at some position $j\geq i$, and $\theta_1$ holds at every position from $i$ up to, but excluding, $j$ \\
$\ltlprev\theta$ & yesterday~\cite{icaps2023bdffgs} & $i\geq1$ and $\theta$ holds at $i-1$; false at position $0$ \\
$\theta_1\ltlsince\theta_2$ & since & $\theta_2$ holds at some position $j\leq i$, and $\theta_1$ holds at every position after $j$ up to and including $i$ \\
\midrule
$\ltleventually\theta\equiv\top\ltluntil\theta$ & sometime & $\theta$ holds at $i$ or at a later position \\
$\ltlalways\theta\equiv\neg\ltleventually\neg\theta$ & always & $\theta$ holds at every position from $i$ to the last \\
$\ltlonce\theta\equiv\top\ltlsince\theta$ & once & $\theta$ holds at $i$ or at an earlier position \\
\bottomrule
\end{tabular}
\caption{The temporal operators of the trajectory constraints of \Cref{def:planning-problem}.}
\label{tab:ltl-operators}
\end{table}

\textbf{The Shell planning problem $\planningtask$.}
The domain declares object types, fluents as predicates and a function, and five action schemas: three name their actor first and, when the action targets another object, their target second, and two are happenings. 
\Cref{lst:shell-pddl} lists the planning problem modelling the Shell example used in this paper.
Six Boolean fluents record whether the cartel is coordinated ($\mathit{coordinated}$), whether the region is at war ($\mathit{war}$), whether an embargo against the importer is in force ($\mathit{embargoed}$), whether the importer has resupplied its ally ($\mathit{resupplied}$), whether the importer and its ally are allied ($\mathit{allied}$), and whether the ally and its neighbour are hostile ($\mathit{hostile}$).
The fluent $\mathit{capacity}$ records the importer's capacity to withstand economic harm, with the totally ordered range $\{\mathit{low},\mathit{normal},\mathit{high}\}$.
The problem names the objects, the initial state and the empty goal.
The goal is $\top$, every action costs $1$, and the exercise budget is $\costthreshold$ (i.e., a narrative holds at most some number of events).

\begin{lstlisting}[morecomment={[l]{;}},caption={Shell's exercise in PDDL.},label={lst:shell-pddl}]
(define (domain shell-1972)
  (:requirements :strips :typing :negative-preconditions :numeric-fluents
                 :constraints :preferences)
  (:types producer consumer country)        ; the cartel, the importer, the states
  (:predicates (coordinated) (war) (embargoed) (resupplied)
               (allied ?i - consumer ?a - country) (hostile ?a ?b - country))
  (:functions (capacity))                   ; 0 low, 1 normal, 2 high
  (:action outbreak :parameters (?a ?b - country)  ; a happening: no actor
    :precondition (and (hostile ?a ?b) (not (war))) :effect (war))
  (:action resupply :parameters (?i - consumer ?a - country)
    :precondition (and (allied ?i ?a) (war)) :effect (resupplied))
  (:action embargo :parameters (?c - producer ?i - consumer)
    :precondition (and (coordinated) (resupplied) (not (embargoed))) ; no second embargo
    :effect (and (embargoed) (decrease (capacity) 1)))
  (:action split :parameters ()                   ; a happening: no actor
    :precondition (coordinated) :effect (not (coordinated)))
  (:action growth :parameters (?i - consumer)
    :precondition (not (coordinated)) :effect (increase (capacity) 1))
  ; the invariant of the subject, a trajectory constraint
  (:constraints (always (imply (embargoed) (coordinated)))))

(define (problem shell-1972-before-the-war) (:domain shell-1972)
  (:objects cartel - producer importer - consumer ally neighbour - country)
  (:init (coordinated) (allied importer ally) (hostile ally neighbour) (= (capacity) 1))
  (:goal (and))                             ; no goal: a narrative explores what can happen
  ; the actors' objectives, one metric each, every preference at weight 1:
  ;   importer: maximize (+ (capacity) (- 1 (is-violated importer-resupplied)))
  ;   cartel:   maximize (- 1 (is-violated cartel-embargoed))
  (:constraints (and (preference importer-resupplied (at end (resupplied)))
                     (preference cartel-embargoed    (at end (embargoed))))))
\end{lstlisting}

\begin{definition}[Plan]\label{def:plans}
 Given a planning problem $\planningtask$, a \emph{plan} $\singleplan{}=\langle\action{1},\action{2},\ldots,\action{m}\rangle$ is a sequence of actions such that each $\action{i}\in\allaction$, all transitions occurring in the sequence are in $\transitionfn$, executing the sequence from $\initialstate$ yields a state satisfying $\goalstate$ (i.e., $\transitionfn(\ldots\transitionfn(\transitionfn(\initialstate,\action{1}),\action{2})\ldots,\action{m})\models\goalstate$, which we write $\transitionfn(\initialstate,\singleplan{})\models\goalstate$), and the plan satisfies every trajectory constraint (i.e., $\singleplan{}\models\theta$ for every $\theta\in\trajconstraints$).
 Its \emph{final state} is $\transitionfn(\initialstate,\singleplan{})$, and its cost is the sum of its action costs, $\costfn(\singleplan{})=\sum_{i=1}^{m}\costfn(\action{i})$; for simplicity, we overload the notation $\costfn$.
 We denote the set of all plans for $\planningtask$ as $\allplans{}$ (i.e., $\allplans{}=\{\singleplan{}\ \vert\ \singleplan{}\text{ is a plan for }\planningtask\}$).
\end{definition}

\textbf{Shell narratives as plans.} The disruption narrative is $\singleplan{d}=\langle\mathit{outbreak},\mathit{resupply},\mathit{embargo}\rangle$, and the continuity narrative is $\singleplan{c}=\langle\mathit{split},\mathit{growth}\rangle$, in which the cartel splits, the region stays at peace, and the economy grows.

In scenario planning, we pair the planning problem $\planningtask$ with the inputs that the scenario criteria read, and we call the pair a \emph{topic} $\topic$, which takes the form of the inputs to a narrative planning problem~\cite{riedl2010narrative}. Narrative planning plans a story for an author: a planning problem whose actions are performed by characters, each with its own goals, and whose plans must reach the author's goal while each character acts in its own interest. \citet{ware2021sabre} assigns each character a numeric utility instead of goals. We use the same inputs, with the actors as the characters and their objectives as the utilities, and we replace the author's goal with the scenario criteria.
The analyst writes these utilities using another PDDL3 feature, \emph{preference} (i.e., a soft constraint a plan may violate at a price): a preference names a condition on the final state, and the problem's metric charges each violated preference a weight.
We give each actor its own metric, so an actor's objective is the weighted sum of the preferences it holds at a state, and an action serves the actor when it raises that sum. \Cref{sec:engine:plausibility} defines the objective and states how the engine reads it.
PDDL3 allows one metric per problem, so the preferences appear in the problem file, with the actors' metrics beside them, as \Cref{lst:shell-pddl} shows.
The problem posed over a topic is then the scenario planning problem, which asks for a set of narratives. We define a topic as follows.

\begin{definition}[Topic]\label{def:topic}
 A \emph{topic} is a tuple $\topic=\langle\planningtask,\actors,\objectives,\concernset,\driverset,\expectedoutcome\rangle$, where $\planningtask$ is a planning problem (\Cref{def:planning-problem}); $\actors$ is the set of \emph{actors} (the characters), $\objectives$ are the actors' objectives (the characters' utilities), one $\objectiveof{\anactor}$ per actor $\anactor\in\actors$; and $\concernset$ is the set of concern fluents, $\driverset$ the list of drivers and $\expectedoutcome$ the conventional outcome, which the paragraphs below define.
\end{definition}

From now on, we refer to a $\planningtask$ plan as a \emph{narrative} (a sequence of permitted events), while we reserve the term plan for the general planning problem of \Cref{def:diversity-problem} and the planners that address it.
A narrative should not be confused with a scenario, because a scenario must also meet the four criteria of \Cref{sec:background:scenario-planning}.
The analyst bounds a narrative's cost with a \emph{budget} $\costthreshold$ (i.e., the highest cost a narrative may have).
In this work, every action costs $1$, so a narrative's cost is its number of actions; we leave actions with different costs for future work.
The budget never ranks one narrative above another.
We can map the elements of a scenario exercise onto the definition of a planning problem and onto the other inputs of a topic as follows.

\paragraph{\textbf{Shell's current situation.}} The current situation is the initial state, denoted $\initialstate$. In Shell's exercise, it is the situation of 1972 that the \texttt{:init} block of \Cref{lst:shell-pddl} records: $\mathit{coordinated}$, $\mathit{allied}$ and $\mathit{hostile}$ are true, $\mathit{war}$, $\mathit{embargoed}$ and $\mathit{resupplied}$ are false, and $\mathit{capacity}$ is $\mathit{normal}$.
Both narratives of \Cref{ex:shell-setting} start from this state.

\paragraph{\textbf{Shell's actors and objectives.}} The participants are the actors $\actors$. The topic lists the actors' objectives $\objectives$ (i.e., each actor's weighted preferences). 
The actors are the cartel, the importer and the ally.
Each is an object of \Cref{lst:shell-pddl}, and each action schema names its actor as its first parameter and, when it is directed at another object, its target as its second: the importer resupplies its ally and grows its economy, the cartel imposes an embargo on the importer, and the ally acts in no schema, so it appears only as the target of the resupply. 
\citet{riedl2010narrative} uses the term happening for actions that no character intends. 
For instance, the war's outbreak and the cartel's split are happenings.
The importer wants its ally resupplied and its capacity high, and the cartel wants the embargo in force, which the two preferences and the two metrics of \Cref{lst:shell-pddl} state.

\paragraph{\textbf{Shell's concern fluents.}} The set $\concernset\subseteq\allfluents$ of concern fluents captures what the organisation cares about (i.e., the fluents that must end with a value other than their initial value for a narrative to be relevant).
Shell's concern is the importer's ability to buy oil, which $\mathit{capacity}$ records, so $\concernset=\{\mathit{capacity}\}$.

\paragraph{\textbf{Shell's drivers.}} The drivers are a list of distinct fluents $\driverset=\langle\fluent_1,\ldots,\fluent_n\rangle$ with each $\fluent_i\in\allfluents$, whose final values distinguish one scenario from another. 
The drivers are the cartel's coordination and the regional war, so $\driverset=\langle\mathit{coordinated},\mathit{war}\rangle$; the disruption narrative ends with $\langle\mathit{true},\mathit{true}\rangle$ and the continuity narrative with $\langle\mathit{false},\mathit{false}\rangle$.

\paragraph{\textbf{Shell's conventional outcome.}} The conventional outcome $\expectedoutcome$ is what the organisation expects of the drivers (i.e., one value for each fluent in $\driverset$).
For Shell, it is the future of a coordinated cartel in a peaceful region (i.e., $\expectedoutcome=\langle\mathit{true},\mathit{false}\rangle$), which the drivers also read in the initial state.
It is one of the two outcomes that Shell's team left unexamined.

\subsection{Planners}\label{sec:background:planners}

In \Cref{sec:background:planning}, we introduced a planning problem whose plans are narratives.
Since a scenario exercise requires multiple narratives, this subsection defines the problem of finding a set of plans and reviews the planners that solve it.
\citet{abdelwahed2026behaviour} states this problem generically, using a \emph{diversity indicator function} (i.e., a function that maps a set of plans to a number, with larger values indicating greater diversity).

\begin{definition}[Diversity planning problem]\label{def:diversity-problem}
Given a planning problem $\planningtask$, a cost bound $\costthreshold$, a number of plans $\planscount$, and a diversity indicator function $\diversityindicator:2^{\allplans}\rightarrow\mathbb{R}$ over sets of plans, a \emph{diversity planning problem} is to find a set of plans $\diversesetplan{\planningtask}\subseteq\allplans$ such that $\costfn(\singleplan{})\leq\costthreshold$ for all $\singleplan{}\in\diversesetplan{\planningtask}$ and $\vert\diversesetplan{\planningtask}\vert\leq\planscount$, and that either maximises $\diversityindicator(\diversesetplan{\planningtask})$ or satisfies $\diversityindicator(\diversesetplan{\planningtask})\geq\distancethreshold$ for a given threshold $\distancethreshold$.
\end{definition}

A solution to a diversity planning problem is a plan set containing at most $\planscount$ plans for $\planningtask$, each within the cost bound, and is either optimal in indicator value or has value at least $\distancethreshold$.
Diverse planners solve this problem.
A diverse planner generates a set of distinct plans according to a diversity model, often a distance function that measures plan similarity.
Early methods relied on domain-specific knowledge: \citet{tate1998generation} with $\plannername{O-Plan}{}{}$, and \citet{myers1999generating} with an HTN planner guided by metatheoretic knowledge.
The first domain-independent method was by \citet{srivastava2007domain}, who modified the constraint-based planner $\plannername{GP-CSP}{}{}$ and the local-search planner $\plannername{LPG}{}{}$ to produce plans with pairwise distances above a threshold.
Other planners varied the distance metric or search strategy, such as $\plannername{DLAMA}{}{}$~\cite{bryce2014landmark}, which uses landmarks, and $\plannername{LPG-d}{}{}$~\cite{nguyen2012generating}, which requires a minimum action distance.
\citet{katz2020reshaping} approached the planning task differently, developing the $\fiplanner{}$ based on \citet{vadlamudi2016combinatorial}'s two-phase method.
The first phase generates a large plan set, and the second selects a diverse subset using clustering or greedy selection.
\citet{katz2022bounding} use mixed-integer programming to find an optimally diverse set, while \citet{zhong2024bi} use beam search based on quality and diversity.
\citet{abdelwahed2026behaviour} introduced the $\fbixoperator{}$, which considers diversity during planning while respecting the cost bound, by forbidding the behaviour of every plan it finds.

Narrative planners use the same inputs as a topic, but none of them is diverse planners, so we cannot use them to generate scenarios. Besides, narrative planners expect a goal state, whereas scenario planning asks which final states can arise. Examples of narrative planners include: 
\citet{riedl2010narrative} search the space of partial plans for a single story in which characters act intentionally;
\citet{haslum2012narrative} reformulates the problem as classical planning, so a classical planner returns one story per call;
\citet{porteous2009controlling} steer a planner toward one story by constraining its trajectory; and
Sabre~\cite{ware2021sabre} searches for one story in which every character acts on its own utility.

In the next section, we describe how our engine generates scenarios for the planning problem in \Cref{sec:background:planning}.
 

%% file: sections-v5/03-engine.tex
\section{Plyscene: A Generic Scenario Planning Engine}\label{sec:engine}

\Cref{sec:background:planning} defined a planning problem whose plans are topic narratives, and \Cref{sec:background:planners} defined the diversity planning problem and the planners that solve it.
This section presents Polyscene, the engine that generates scenarios for any given topic.
The engine takes a topic $\topic$ and a budget $\costthreshold$ and returns one scenario for each driver outcome that a scenario within budget can reach, so the analyst reads one narrative per future rather than a pile of variations on the same future.
The engine is designed to be generic as follows.
The five criteria are formulated and implemented once for every planning problem structure.
This structure includes an action's actor and target, the fluents that define its preconditions, effects, objectives, and drivers.
The engine does not change its rules based on the specific details of a topic. 
Instead, a new topic only changes the facts the engine interprets.

\Cref{sec:background:scenario-planning} gave the criteria that the engine enforces.
These criteria are (1) a scenario is \emph{plausible} when it can happen; (2) \emph{internally consistent} when the logics it combines agree; (3) \emph{relevant} when it bears on the organisation's concerns; (4) \emph{challenging} when it contests the organisation's conventional expectation; and (5) \emph{differentiation} enforces that the scenarios must differ structurally rather than as variations on one base case.
Each criterion that judges one scenario becomes a trajectory constraint (\Cref{def:planning-problem}) that the engine adds to $\trajconstraints$, so the extended planning problem's narratives are exactly the scenarios.
Differentiation compares narratives, so it decides which narratives the engine returns rather than which narratives exist.
We define a scenario as follows.

\begin{definition}[Scenario]\label{def:scenario}
 Given a topic $\topic$ with planning problem $\planningtask$, a \emph{scenario} is a plausible, internally consistent, relevant, and challenging narrative of $\planningtask$ (\Cref{def:plausible,def:consistent,def:relevant,def:challenging}).
 We denote the set of all scenarios of the topic as $\allscenarios\subseteq\allplans$.
\end{definition}

\Cref{sec:engine:criteria} formulates each criterion as a trajectory constraint, uses Shell's exercise as a running example, and defines the scenario-planning problem.
\Cref{sec:engine:sat} gives the planning-as-satisfiability query that carries the constraints, and \Cref{sec:engine:asp} implements each constraint in Answer Set Programming (ASP)~\cite{gebser2012answer}.

\subsection{The Criteria as Trajectory Constraints}\label{sec:engine:criteria}

Each subsection covers a criterion regarding the scenario planning literature, applies it to Shell's exercise, defines it, and provides its trajectory constraint $\critformula{x}$.

\subsubsection{Internal Consistency}\label{sec:engine:consistency}
A scenario is \emph{internally consistent} when it contains no inherent contradictions or inconsistencies~\cite{wilson1998mental}.
\citet{schoemaker1995scenario} evaluates this consistency by asking whether the scenario's different outcomes can realistically occur together.
Which values can occur together is a question about the subject, so the analyst answers it when writing $\planningtask$, with two patterns that \Cref{lst:shell-pddl} shows.
The first is an \emph{invariant} (i.e., a formula $\psi$ over $\allfluents$ that must hold in every state), which the analyst writes as the trajectory constraint $\ltlalways\psi$ in $\trajconstraints$ (e.g., an embargo in force requires a coordinated cartel, so $\ltlalways(\mathit{embargoed}\rightarrow\mathit{coordinated})$).
The second is a fluent that records that an action has happened and whose precondition requires it to be false, so the action does not repeat.
For instance, the embargo sets $\mathit{embargoed}$ and requires it to be false, so the cartel imposes no second embargo, much as a country does not mobilise against the same country twice.
The second pattern needs no constraint, since a plan satisfies every precondition, so the criterion reduces to the invariants.
We define internal consistency as follows.

\begin{definition}[Internal consistency]\label{def:consistent}
 Given a topic $\topic$ with planning problem $\planningtask$ and the invariants $\psi_1,\ldots,\psi_l$ of the subject (i.e., the formulas over $\allfluents$ whose constraints $\ltlalways\psi_1,\ldots,\ltlalways\psi_l$ the analyst wrote into $\trajconstraints$), a narrative $\singleplan{}=\langle\action{1},\ldots,\action{m}\rangle$ with the states $\state{0}=\initialstate$ and $\state{i}=\transitionfn(\state{i-1},\action{i})$ is \emph{internally consistent} if every state $\state{i}$ satisfies every invariant $\psi_1,\ldots,\psi_l$.
\end{definition}

The constraint $\critformula{cons}=\bigwedge_{k=1}^{l}\ltlalways\psi_k$ implements \Cref{def:consistent} in LTL$_f$.
In Shell's exercise, $\critformula{cons}=\ltlalways(\mathit{embargoed}\rightarrow\mathit{coordinated})$ rejects $\langle\mathit{outbreak},\mathit{resupply},\mathit{embargo},\mathit{split}\rangle$, whose final state has the embargo in force and the cartel divided.

\subsubsection{Relevance}\label{sec:engine:relevance}

According to~\citet{wilson1998mental}, a scenario is \emph{relevant} when it offers specific insights into the future that aid in decision-making.
We capture this through the topic's concern fluents $\concernset$ (\Cref{def:topic}), because a scenario where concerns remain unchanged offers little insight for decision-making.
We use a concern's final value as a stand-in for insight because decision-makers tend to interpret a scenario by how it ends.
In Shell's exercise, the concern is the importer's capacity (i.e., $\concernset=\{\mathit{capacity}\}$).
The disruption narrative $\singleplan{d}=\langle\mathit{outbreak},\mathit{resupply},\mathit{embargo}\rangle$ concludes with capacity at $\mathit{low}$, as the embargo reduces it, giving Shell insight into its market.
The narrative $\singleplan{r}=\langle\mathit{outbreak},\mathit{resupply}\rangle$ concludes before the embargo and keeps capacity at $\mathit{normal}$.
It provides no information relevant to Shell's decision, even though a war has erupted.

\begin{definition}[Relevance]\label{def:relevant}
 Given a topic $\topic$ with planning problem $\planningtask$ and concern fluents $\concernset\subseteq\allfluents$, a narrative $\singleplan{}$ is \emph{relevant} if $\exists\fluent\in\concernset.\ \transitionfn(\initialstate,\singleplan{})[\fluent]\neq\initialstate[\fluent]$ (i.e., at least one concern fluent ends with a value other than its initial one).
\end{definition}

The constraint $\critformula{rel}=\ltlalways\ltleventually\bigvee_{\fluent\in\concernset}\fluent\neq\initialstate[\fluent]$ implements \Cref{def:relevant} in LTL$_f$, where each $\initialstate[\fluent]$ is a value read from the initial state.
A formula $\ltlalways\ltleventually\psi$ is satisfied if and only if $\psi$ holds at the final position.
In Shell's exercise, the constraint becomes $\ltlalways\ltleventually(\mathit{capacity}\neq\mathit{normal})$.
The disruption narrative satisfies it because it ends with capacity at $\mathit{low}$, whereas $\singleplan{r}$ does not, because it ends with capacity at $\mathit{normal}$.

\subsubsection{Challenge}\label{sec:engine:challenge}

\citet{wilson1998mental} describes a scenario as \emph{challenging} when it contests the organisation's entrenched beliefs regarding the future.
We judge whether a scenario is challenging by the values its final state gives the drivers $\driverset$, which we call its \emph{outcome}: for drivers $\driverset=\langle\fluent_1,\ldots,\fluent_n\rangle$, the outcome of a narrative $\singleplan{}$ is the tuple $\outcome{\driverset}{\singleplan{}}=\langle\transitionfn(\initialstate,\singleplan{})[\fluent_1],\ldots,\transitionfn(\initialstate,\singleplan{})[\fluent_n]\rangle$, and we write $\outcome{\driverset}{\singleplan{}}[\fluent_i]$ for its $i$-th value, as for every tuple over $\driverset$.
A narrative is challenging when its outcome differs from the topic's conventional outcome $\expectedoutcome$ (\Cref{def:topic}).
The conventional outcome is an input in its own right, and it coincides with the initial values of the drivers in Shell's exercise only because Shell expected no change; an organisation that expected the cartel to split would state $\expectedoutcome=\langle\mathit{false},\mathit{false}\rangle$, and the engine would then treat a coordinated cartel as the challenging case.
The disruption narrative $\singleplan{d}=\langle\mathit{outbreak},\mathit{resupply},\mathit{embargo}\rangle$ leads to an outcome where the cartel remains coordinated, but the region is at war (i.e., $\outcome{\driverset}{\singleplan{d}}=\langle\mathit{true},\mathit{true}\rangle$).
Since this outcome differs from $\expectedoutcome=\langle\mathit{true},\mathit{false}\rangle$ on the war driver, the narrative is challenging.

\begin{definition}[Challenge]\label{def:challenging}
 Given a topic $\topic$ with planning problem $\planningtask$, drivers $\driverset$ and conventional outcome $\expectedoutcome$, a narrative $\singleplan{}$ is \emph{challenging} if $\outcome{\driverset}{\singleplan{}}\neq\expectedoutcome$.
\end{definition}

The constraint $\critformula{chal}=\ltlalways\ltleventually\bigvee_{i=1}^{n}\fluent_i\neq\expectedoutcome[\fluent_i]$ implements \Cref{def:challenging} in LTL$_f$, which holds when the final state differs from $\expectedoutcome$ on at least one driver.
In Shell's exercise, $\expectedoutcome=\langle\mathit{true},\mathit{false}\rangle$ gives $\critformula{chal}=\ltlalways\ltleventually(\neg\mathit{coordinated}\vee\mathit{war})$.
The disruption narrative satisfies it, since it ends in war.

\subsubsection{Plausibility}\label{sec:engine:plausibility}

\begin{table}[t]
\centering
\footnotesize
\renewcommand{\arraystretch}{1.15}
\begin{tabular}{@{}l>{\raggedright\arraybackslash}p{2.6cm}>{\raggedright\arraybackslash}p{6.2cm}>{\raggedright\arraybackslash}p{3.1cm}@{}}
\toprule
Condition & Judges & A narrative breaks it when & Source \\
\midrule
(1) intentional & an actor's action & the action raises no objective of its actor, replies to no action aimed at its actor, and starts no causal chain that ends in an action that does either & intentionality~\cite{riedl2010narrative}, read through utilities~\cite{ware2021sabre} \\
(2) connected & an actor's action after the first & no causal link reaches it from an earlier action, and it replies to none and undoes none & causal links~\cite{mcallester1991systematic} \\
(3) connected & a happening & no causal link leads from it to a later action of an actor & happenings~\cite{riedl2010narrative} \\
(4) loop-free & an actor's action & the state before the action occurred earlier in the narrative & loop-free plans~\cite{kautz1992planning} \\
\bottomrule
\end{tabular}
\caption{The plausibility conditions and their sources.}
\label{tab:plausibility-laws}
\end{table}

\citet{wilson1998mental} calls a scenario \emph{plausible} if it can occur.
We read this as two conditions on every narrative action: the action can be executed in its state, and it makes sense there.
The first condition holds by default, since every narrative is a $\planningtask$ plan.
The second requires additional constraints, since a precondition lets the cartel impose an embargo but does not explain why.
We borrow the two that narrative planning uses to make a story believable and the one that bounded planning uses to keep a plan loop-free.
Narrative planning requires every character action to be \emph{intentional} (i.e., a step of a plan towards one of the character's goals) and every event to be causally connected to the rest of the story~\cite{riedl2010narrative}. Sabre~\cite{ware2021sabre} reads intentions through a numeric utility per character, so that an action is intentional when it begins a plan that raises its actor's utility.
We adopt this approach, using actors' objectives as utilities and the causal links of partial-order planning~\cite{mcallester1991systematic} as connections. 
We define both below.

\Cref{tab:plausibility-laws} lists the three resulting conditions and their sources.
Conditions are only as effective as the objectives the analyst defines. 
A condition will never trigger if the topic leaves out the input it depends on (for example, an intentional condition in a topic with no objectives). 

\paragraph{\textbf{Objectives and causal links.}}
The conditions read the actors' objectives $\objectives$ in $\topic$ and the causal link. We define both before stating the conditions.

\emph{Objectives.}
Each actor $\anactor$ has an \emph{objective} $\objectiveof{\anactor}:\allstate\rightarrow\mathbb{R}$ (i.e., a function that assigns a number to each state), which is the utility that Sabre gives each character~\cite{ware2021sabre}.
Sabre allows any numeric expression over the state. 
We restrict the objective to a weighted sum so a PDDL3 metric can express it.
The analyst writes such objectives as PDDL3 preferences~\cite{gerevini2009deterministic}, each with a weight: a preference $\langle \anactor,\fluent,\avalue,\aweight\rangle\in\objectives$ states that actor $\anactor$'s objective gains $\aweight>0$ while the fluent $\fluent\in\allfluents$ holds the value $\avalue\in\fluentdomain{\fluent}$. 
For a numeric fluent, $\avalue$ is a direction, $\mathit{up}$ or $\mathit{down}$, and the objective gains or loses $\aweight$ per unit of $\fluent$.
PDDL3 metric \texttt{is-violated} computes the objective of $\anactor$ at a state $\state{}$ as sum of the weights $\aweight$ of its preferences $\langle\anactor,\fluent,\avalue,\aweight\rangle$ with $\state{}[\fluent]=\avalue$, plus $\aweight\cdot\state{}[\fluent]$ for each numeric preference $\langle\anactor,\fluent,\mathit{up},\aweight\rangle$ and minus $\aweight\cdot\state{}[\fluent]$ for each $\langle\anactor,\fluent,\mathit{down},\aweight\rangle$. 
An action \emph{raises} $\anactor$'s objective when the state it reaches has a higher $\objectiveof{\anactor}$ than the state before it.
For instance, \Cref{lst:shell-pddl} gives the importer two preferences and the cartel one.
The preference \texttt{importer-resupplied} is $\langle\mathit{importer},\mathit{resupplied},\mathit{true},1\rangle$: the importer's objective gains $1$ while $\mathit{resupplied}$ is true.
The term \texttt{(capacity)} of the importer's metric is $\langle\mathit{importer},\mathit{capacity},\mathit{up},1\rangle$: the objective gains $1$ per unit of capacity.
The preference \texttt{cartel-embargoed} is $\langle\mathit{cartel},\mathit{embargoed},\mathit{true},1\rangle$.
Along the disruption narrative, the importer's objective is $1$ in the initial state (the capacity), still $1$ after the outbreak, $2$ after the resupply, and $1$ after the embargo, which lowers the capacity to $0$ while $\mathit{resupplied}$ stays true; the cartel's objective is $0$ until the embargo raises it to $1$.
So the resupply raises the importer's objective and the embargo raises the cartel's.

\emph{Causal links.}
An action that raises no objective can still matter because of what it prepares or what others did, the causal connectedness readers look for in a story~\cite{trabasso1985causal}.
A \emph{causal link}~\cite{mcallester1991systematic} goes from an action to a later one when the first changes a fluent that the second requires as a precondition, and no action between them changes that fluent. 
A \emph{causal chain} is a sequence of causal links, of any length within the budget.
In \Cref{lst:shell-pddl}, the outbreak sets $\mathit{war}$, enabling resupply; resupply sets $\mathit{resupplied}$, enabling embargo. Thus, outbreak causally enables resupply, and resupply causally enables embargo, forming a causal chain among the three actions.
Actions relate to their actors and targets in two key ways. 
An action \emph{replies} to an earlier action when it is directed at the same actor as that earlier action. 
An action \emph{undoes} a previous action by another actor when it changes a result back to a value that the earlier action set, and that value has remained the same since.
In Shell's exercise, no action replies to or undoes another. This is because no two actors direct actions at each other, and no action reverses another's effect.

\paragraph{\textbf{The conditions.}}
We now state the three plausibility conditions (i.e., intentional, connected, and loop-free) and check each against the disruption narrative $\singleplan{d}=\langle\mathit{outbreak},\mathit{resupply},\mathit{embargo}\rangle$, which is plausible.

The \emph{intentional} condition asks that an actor's action serves its actor (i.e., the action raises its actor's objective, replies to an action aimed at its actor, or starts a causal chain that ends in an action that raises its actor's objective or in a reply of its actor's own). 
This can also be seen as setting up a motivated action.
A chain may pass through other actors' actions, since an actor may act so that another acts, and it may have any length within the budget.
In $\singleplan{d}$, the resupply raises the importer's objective, since it sets $\mathit{resupplied}$, which the importer wants, and the embargo raises the cartel's, since it sets $\mathit{embargoed}$, so both are intentional.
The embargo lowers the importer's capacity, which costs the importer, but the condition judges each action by its own actor's objective.
Note that, if the topic did not provide any cartel preference, the intentional condition would have rejected $\singleplan{d}$ because the embargo starts no chain and replies to nothing.

The \emph{connected} condition requires the narrative to be one story.
Every actor's action after the first causally links to an earlier action, replies to one, or undoes one, so the narrative doesn't split into unrelated threads. 
Every happening causally links to a later actor's action.
Thus, no event happens without consequence.
In $\singleplan{d}$, the resupply is the first actor's action, which the condition exempts; a causal link leads from the resupply to the embargo, since the embargo reads $\mathit{resupplied}$. 
A happening action leads from the outbreak to the resupply, since the resupply reads $\mathit{war}$, so both conditions hold.

The \emph{loop-free} condition prevents an actor from acting from a state the narrative has already passed through, since intervening actions would lead nowhere.
In $\singleplan{d}$, no state repeats.

\begin{table}[!tp]
\centering
\footnotesize
\renewcommand{\arraystretch}{1.2}
\begin{tabular}{@{}>{\raggedright\arraybackslash}p{6.4cm}>{\raggedright\arraybackslash}p{5.7cm}>{\raggedright\arraybackslash}p{3.3cm}@{}}
\toprule
\multicolumn{3}{@{}p{16.3cm}@{}}{Given a narrative $\singleplan{}=\langle\action{1},\ldots,\action{m}\rangle$ with the states $\state{0}=\initialstate$ and $\state{i}=\transitionfn(\state{i-1},\action{i})$, an actor $\anactor\in\actors$, a fluent $\fluent\in\allfluents$, a value $\avalue\in\fluentdomain{\fluent}$, positions $i,j,k\in\{1,\ldots,m\}$, and a bound $h$ on the number of actions of a narrative; $\mathrm{vars}(\actionpre_{\action{}})$ denotes the fluents that occur in the precondition of $\action{}$, $\happenings\subseteq\allaction$ the happenings, $\allstates$ the states of $\planningtask$, $\hat{\state{}}=\bigwedge_{\fluent\in\allfluents}\fluent=\state{}[\fluent]$ the formula that holds in the state $\state{}$ alone, and $c$ ranges over the values that an objective takes.} \\
\midrule
First-order definition & LTL$_f$ implementation & Holds when \\
\midrule
$\mathrm{acts}(\action{i},\anactor):=\ \actorof{\action{i}}=\anactor$ & $\mathrm{Acts}_{\anactor}:=\bigvee\{\action{}\in\allaction\mid\actorof{\action{}}=\anactor\}$ & the action $\action{i}$ is the actor $\anactor$'s \\
$\mathrm{actorActs}(\action{i}):=\ \exists\anactor\in\actors:\ \mathrm{acts}(\action{i},\anactor)$ & $\mathrm{ActorActs}:=\bigvee_{\anactor\in\actors}\mathrm{Acts}_{\anactor}$ & the action $\action{i}$ has an actor \\
$\mathrm{hap}(\action{i}):=\ \neg\mathrm{actorActs}(\action{i})$ & $\mathrm{Hap}:=\bigvee\happenings$ & the action $\action{i}$ is a happening \\
$\mathrm{reads}(\action{i},\fluent):=\ \fluent\in\mathrm{vars}(\actionpre_{\action{i}})$ & $\mathrm{Reads}_{\fluent}:=\bigvee\{\action{}\in\allaction\mid\fluent\in\mathrm{vars}(\actionpre_{\action{}})\}$ & the action $\action{i}$ reads $\fluent$ \\
\midrule
$\mathrm{repeated}(\state{i}):=\ \exists j<i:\ \state{j}=\state{i}$ & $\mathrm{Rep}:=\bigvee_{\state{}\in\allstates}\big(\hat{\state{}}\wedge\ltlprev\ltlonce\hat{\state{}}\big)$ & the state $\state{i}$ occurred at an earlier position \\
$\mathrm{chg}(\action{i},\fluent):=\ \state{i}[\fluent]\neq\state{i-1}[\fluent]$ & $\mathrm{Chg}_{\fluent}:=\bigvee_{\avalue\in\fluentdomain{\fluent}}\big(\fluent=\avalue\wedge\ltlprev\neg(\fluent=\avalue)\big)$ & the action $\action{i}$ changes the value of $\fluent$ \\
$\mathrm{gave}(\action{i},\fluent,\avalue):=\ \mathrm{chg}(\action{i},\fluent)\wedge\state{i}[\fluent]=\avalue$ & $\mathrm{Gave}_{\fluent,\avalue}:=\mathrm{Chg}_{\fluent}\wedge\fluent=\avalue$ & the action $\action{i}$ gives $\fluent$ the value $\avalue$ \\
$\mathrm{raises}(\action{i},\anactor):=\ \objectiveof{\anactor}(\state{i})>\objectiveof{\anactor}(\state{i-1})$ & $\mathrm{Raises}_{\anactor}:=\bigvee_{c}\big(\objectiveof{\anactor}=c\wedge\ltlprev(\objectiveof{\anactor}<c)\big)$ & the action $\action{i}$ raises the objective of $\anactor$ \\
\midrule
$\mathrm{link}(\action{i},\action{j}):=\ i<j\wedge\exists\fluent:\ \mathrm{reads}(\action{j},\fluent)\wedge\mathrm{chg}(\action{i},\fluent)\wedge\forall k,\ i<k<j:\ \neg\mathrm{chg}(\action{k},\fluent)$ & $\mathrm{Linked}:=\bigvee_{\fluent\in\allfluents}\big(\mathrm{Reads}_{\fluent}\wedge\ltlprev(\neg\mathrm{Chg}_{\fluent}\ltlsince\mathrm{Chg}_{\fluent})\big)$ \newline $\mathrm{LinksTo}(\theta):=\bigvee_{\fluent\in\allfluents}\big(\mathrm{Chg}_{\fluent}\wedge\ltlnext(\neg\mathrm{Chg}_{\fluent}\ltluntil(\mathrm{Reads}_{\fluent}\wedge\theta))\big)$ & a causal link leads from $\action{i}$ to $\action{j}$: the first changes a fluent that the second reads, and nothing between them changes it; Linked is read at $\action{j}$, and LinksTo($\theta$) at $\action{i}$ with $\theta$ required of $\action{j}$ \\
$\mathrm{replies}(\action{i}):=\ \exists j<i:\ \actorof{\action{j}}=\targetof{\action{i}}\wedge\targetof{\action{j}}=\actorof{\action{i}}$ & $\mathrm{Replies}:=\bigvee_{\action{}\in\allaction}\big(\action{}\wedge\ltlprev\ltlonce\bigvee\{\action{}'\in\allaction\mid\actorof{\action{}'}=\targetof{\action{}}\wedge\targetof{\action{}'}=\actorof{\action{}}\}\big)$ & the action $\action{i}$ replies to an earlier action directed at its actor \\
$\mathrm{undoes}(\action{i}):=\ \exists j<i,\ \fluent,\ \avalue:\ \mathrm{gave}(\action{j},\fluent,\avalue)\wedge\mathrm{actorActs}(\action{j})\wedge\actorof{\action{j}}\neq\actorof{\action{i}}\wedge\mathrm{chg}(\action{i},\fluent)\wedge\forall k,\ j\leq k<i:\ \state{k}[\fluent]=\avalue$ & $\mathrm{Undoes}:=\bigvee_{\action{}\in\allaction}\Big(\action{}\wedge\bigvee_{\fluent,\avalue}\big(\mathrm{Chg}_{\fluent}\wedge\ltlprev(\fluent=\avalue\ \ltlsince\ (\mathrm{Gave}_{\fluent,\avalue}\wedge\bigvee\{\action{}'\in\allaction\setminus\happenings\mid\actorof{\action{}'}\neq\actorof{\action{}}\}))\big)\Big)$ & the action $\action{i}$ changes $\fluent$ away from a value that another actor's action $\action{j}$ gave it and that has held since \\
$\mathrm{chain}^{1}(\action{i},\anactor):=\ \mathrm{raises}(\action{i},\anactor)\vee(\mathrm{acts}(\action{i},\anactor)\wedge\mathrm{replies}(\action{i}))$ & $\mathrm{Chain}^{1}_{\anactor}:=\mathrm{Raises}_{\anactor}\vee(\mathrm{Acts}_{\anactor}\wedge\mathrm{Replies})$ & the action $\action{i}$ itself raises the objective of $\anactor$ or is a reply of $\anactor$'s own \\
$\mathrm{chain}^{k+1}(\action{i},\anactor):=\ \mathrm{chain}^{1}(\action{i},\anactor)\vee\exists j:\ \mathrm{link}(\action{i},\action{j})\wedge\mathrm{actorActs}(\action{j})\wedge\mathrm{chain}^{k}(\action{j},\anactor)$ & $\mathrm{Chain}^{k+1}_{\anactor}:=\mathrm{Chain}^{1}_{\anactor}\vee\mathrm{LinksTo}(\mathrm{ActorActs}\wedge\mathrm{Chain}^{k}_{\anactor})$ & a causal chain of at most $k$ links from $\action{i}$ through actors' actions ends in such an action \\
$\mathrm{chain}(\action{i},\anactor):=\ \mathrm{chain}^{m}(\action{i},\anactor)$ & $\mathrm{Chain}_{\anactor}:=\mathrm{Chain}^{h}_{\anactor}$ & the chain may use every link the narrative has, since $m$ actions hold at most $m-1$ links \\
\midrule
$\critformula{int}:=\ \forall i,\ \anactor\in\actors:\ \mathrm{acts}(\action{i},\anactor)\rightarrow\mathrm{chain}(\action{i},\anactor)$ & \multicolumn{2}{>{\raggedright\arraybackslash}p{9.4cm}@{}}{$\ltlalways\bigwedge_{\anactor\in\actors}\big(\mathrm{Acts}_{\anactor}\rightarrow\mathrm{Chain}_{\anactor}\big)$} \\
$\critformula{con}:=\ \forall i:\ \big(\mathrm{actorActs}(\action{i})\wedge\exists j<i:\ \mathrm{actorActs}(\action{j})\big)\rightarrow\big(\exists j<i:\ \mathrm{link}(\action{j},\action{i})\big)\vee\mathrm{replies}(\action{i})\vee\mathrm{undoes}(\action{i})$ & \multicolumn{2}{>{\raggedright\arraybackslash}p{9.4cm}@{}}{$\ltlalways\big((\mathrm{ActorActs}\wedge\ltlprev\ltlonce\mathrm{ActorActs})\rightarrow\mathrm{Linked}\vee\mathrm{Replies}\vee\mathrm{Undoes}\big)$} \\
$\critformula{loop}:=\ \forall i:\ \mathrm{actorActs}(\action{i})\rightarrow\neg\mathrm{repeated}(\state{i-1})$ & \multicolumn{2}{>{\raggedright\arraybackslash}p{9.4cm}@{}}{$\ltlalways\big(\mathrm{ActorActs}\rightarrow\neg\ltlprev\mathrm{Rep}\big)$} \\
$\critformula{hap}:=\ \forall i:\ \mathrm{hap}(\action{i})\rightarrow\exists j:\ \mathrm{link}(\action{i},\action{j})\wedge\mathrm{actorActs}(\action{j})$ & \multicolumn{2}{>{\raggedright\arraybackslash}p{9.4cm}@{}}{$\ltlalways\big(\mathrm{Hap}\rightarrow\mathrm{LinksTo}(\mathrm{ActorActs})\big)$} \\
\bottomrule
\end{tabular}
\caption{The relations over narrative positions required by \Cref{def:plausible}. Implementations are evaluated at positions under the grammar of \Cref{def:planning-problem}, where $\fluent=\avalue$ is a formula over $\allfluents$, and each action atom is in $\allaction$.}
\label{tab:plausibility-formulas} 
\end{table}

\Cref{tab:plausibility-formulas} states the relations that the conditions read, as first-order relations over the positions of a narrative along with its LTL$_f$ implementation.
There are three types of conditions: (1) four types depend only on the action itself (i.e., $\mathrm{acts}$, $\mathrm{actorActs}$, $\mathrm{hap}$, and $\mathrm{reads}$); (2) four types compare the values of fluents across different positions (i.e., $\mathrm{repeated}$, $\mathrm{chg}$, $\mathrm{gave}$, and $\mathrm{raises}$); and (3) four types combine both aspects (i.e., $\mathrm{link}$, $\mathrm{replies}$, $\mathrm{undoes}$, and $\mathrm{chain}$).
The family of actions called $\mathrm{chain}^{k}$ allows for varying numbers of links. In $\mathrm{chain}^{k+1}$, the engine can have up to $k$ links. 
In a narrative involving $m$ events, there can be at most $m-1$ links, so $\mathrm{chain}=\mathrm{chain}^{m}$.
Under these conditions, we define plausibility as follows.

\begin{definition}[Plausibility]\label{def:plausible}
Given a topic $\topic$ with planning problem $\planningtask$ and objectives $\objectives$, a narrative $\singleplan{}=\langle\action{1},\ldots,\action{m}\rangle$ with the states $\state{0}=\initialstate$ and $\state{i}=\transitionfn(\state{i-1},\action{i})$ is \emph{plausible} if it satisfies the following four conditions, stated over the relations of \Cref{tab:plausibility-formulas} with positions $i,j\in\{1,\ldots,m\}$:
\begin{enumerate}
    \item (intentional) $\forall i,\ \anactor\in\actors:\ \mathrm{acts}(\action{i},\anactor)\rightarrow\mathrm{chain}(\action{i},\anactor)$, i.e., every actor's action starts a causal chain that ends in an action raising its actor's objective or in its actor's own reply;
    \item (connected) $\forall i:\ \big(\mathrm{actorActs}(\action{i})\wedge\exists j<i:\ \mathrm{actorActs}(\action{j})\big)\rightarrow\big(\exists j<i:\ \mathrm{link}(\action{j},\action{i})\big)\vee\mathrm{replies}(\action{i})\vee\mathrm{undoes}(\action{i})$, i.e., every actor's action after the first is reached by a causal link, replies to or undoes an earlier action;
    \item (loop-free) $\forall i:\ \mathrm{actorActs}(\action{i})\rightarrow\neg\mathrm{repeated}(\state{i-1})$, i.e., no actor acts from a state that the narrative passed through earlier;
    \item (happening) $\forall i:\ \mathrm{hap}(\action{i})\rightarrow\exists j:\ \mathrm{link}(\action{i},\action{j})\wedge\mathrm{actorActs}(\action{j})$, i.e., every happening leads by a causal link to a later action of an actor.
\end{enumerate}
\end{definition}

\paragraph{\textbf{The trajectory constraint.}}
Plausibility adds the constraint $\critformula{plau}=\critformula{int}\wedge\critformula{con}\wedge\critformula{loop}\wedge\critformula{hap}$ to the trajectory constraints $\trajconstraints$, one formula per condition of \Cref{def:plausible}.
\Cref{tab:plausibility-formulas} states the implementation in LTL$_f$.
In Shell's exercise, the disruption narrative gives $\objectiveof{\mathit{importer}}(\state{2})=2>\objectiveof{\mathit{importer}}(\state{1})=1$, so $\mathrm{raises}(\action{2},\mathit{importer})$ holds, and $\objectiveof{\mathit{cartel}}(\state{3})=1>\objectiveof{\mathit{cartel}}(\state{2})=0$, so $\mathrm{raises}(\action{3},\mathit{cartel})$ holds, and $\critformula{int}$ is satisfied.
The resupply at $2$ changes $\mathit{resupplied}$, which the embargo at $3$ reads, so $\mathrm{link}(\action{2},\action{3})$ holds and $\critformula{con}$ is satisfied; no state repeats, so $\critformula{loop}$ is satisfied.
For the outbreak at $1$, $\critformula{hap}$ asks for a later actor's action $\action{j}$ with $\mathrm{link}(\action{1},\action{j})$: the outbreak changes $\mathit{war}$, and only the resupply reads it, so $\mathrm{link}(\action{1},\action{2})$ holds here, and every narrative in which no resupply follows the outbreak fails $\critformula{hap}$.
\Cref{lst:shell-criteria} states this case for Shell as a PDDL3 constraint.

\subsubsection{Differentiation and the Scenario Planning Problem}\label{sec:engine:differentiation}

The previous parts covered the criteria for a single scenario.
This part defines the criteria for a set of scenarios and the scenario planning problem.
According to \citet{types-scenario-planning}, a differentiated scenario set must differ systematically on key variables, allowing analysts to trace each scenario to a different combination of driver values.
Note that differentiation should not be confused with relevance, because relevance assesses the concern fluents of a single scenario, while differentiation compares the drivers across multiple scenarios.
To clarify this, consider Shell's exercise with the budget $\costthreshold=4$ and two scenarios, $\singleplan{1}=\langle\mathit{outbreak},\mathit{split},\mathit{resupply},\mathit{growth}\rangle$ and $\singleplan{2}=\langle\mathit{split},\mathit{outbreak},\mathit{resupply},\mathit{growth}\rangle$, together with the disruption narrative $\singleplan{d}=\langle\mathit{outbreak},\mathit{resupply},\mathit{embargo}\rangle$.
The scenarios $\singleplan{1}$ and $\singleplan{2}$ differ only in the order of their events.
In $\singleplan{1}$, the war breaks out before the cartel splits, while in $\singleplan{2}$, the cartel splits before the war.
Both scenarios leave a divided cartel in a war-torn region, whereas $\singleplan{d}$ ends with a coordinated cartel in the same region.
Thus, two narratives that reach the same outcome by different routes, or that differ only on fluents outside $\driverset$, count as the same.
This makes the criterion checkable and separates it from SPA's clustering (covered in \Cref{sec:related-work}) because the analyst's drivers fix the grouping before planning rather than after.

\begin{definition}[Differentiation]\label{def:differentiated}
 Given a topic $\topic$ with planning problem $\planningtask$ and drivers $\driverset$, a set of scenarios $\scenarioset\subseteq\allscenarios$ is \emph{differentiated} if $\vert\{\outcome{\driverset}{\singleplan{}}\mid\singleplan{}\in\scenarioset\}\vert=\vert\scenarioset\vert$ (i.e., no two scenarios in $\scenarioset$ share an outcome).
\end{definition}

Differentiation compares outcomes across scenarios, so it conditions on a set of scenarios rather than a formula over one narrative.
For a set of scenarios $\scenarioset'$, the formula $\critformula{diff}(\scenarioset')=\bigwedge_{\singleplan{}'\in\scenarioset'}\ltlalways\ltleventually\bigvee_{i=1}^{n}\fluent_i\neq\outcome{\driverset}{\singleplan{}'}[\fluent_i]$ holds on a narrative whose outcome differs from the outcome of every member of $\scenarioset'$.
A set $\scenarioset$ is differentiated if and only if each of its members satisfies the formula of the others (i.e., $\singleplan{}\models\critformula{diff}(\scenarioset\setminus\{\singleplan{}\})$ for every $\singleplan{}\in\scenarioset$), so every pair of members is compared.
Unlike the other criteria, this condition cannot be checked on a single narrative alone, because each member's formula depends on the others.
In Shell's exercise, the set $\{\singleplan{d},\singleplan{1}\}$ is differentiated.
The disruption narrative satisfies $\critformula{diff}(\{\singleplan{1}\})=\ltlalways\ltleventually(\mathit{coordinated}\vee\neg\mathit{war})$, since it ends with a coordinated cartel, and $\singleplan{1}$ satisfies $\critformula{diff}(\{\singleplan{d}\})=\ltlalways\ltleventually(\neg\mathit{coordinated}\vee\neg\mathit{war})$, since it ends with a divided cartel.
Adding $\singleplan{2}$ breaks the condition: $\critformula{diff}(\{\singleplan{d},\singleplan{1}\})$ requires $\singleplan{2}$ to differ from $\singleplan{1}$, but both end with a divided cartel at war.

We define the scenario planning problem as follows.

\begin{definition}[Scenario planning problem]\label{def:scenario-problem}
 Given a topic $\topic$ with planning problem $\planningtask$ and a budget $\costthreshold$, a \emph{scenario planning problem} is to find a set of scenarios $\scenarioset\subseteq\allscenarios$ with $\costfn(\singleplan{})\leq\costthreshold$ for every $\singleplan{}\in\scenarioset$ that is differentiated and $\vert\scenarioset\vert$ is maximuised.
\end{definition}

The scenario planning problem can be cast as a diversity planning problem (\Cref{def:diversity-problem}) with three design decisions.
First, the planning problem is $\planningtask$ with the goal $\goalstate=\top$ and one more trajectory constraint for each criterion that judges a single scenario (i.e., $\trajconstraints\ =\{\critformula{plau},\critformula{cons},\critformula{rel},\critformula{chal}\}$); its plans are then exactly the scenarios $\allscenarios$.
Second, the cost bound is the budget $\costthreshold$, and the number of plans is $\planscount=\prod_{\fluent\in\driverset}\vert\fluentdomain{\fluent}\vert$ (i.e., the number of possible outcomes), which no differentiated set can exceed, so this bound never removes a solution.
Third, the diversity indicator counts the distinct outcomes of a differentiated set and gives $0$ to any other set (i.e., $\diversityindicator(\scenarioset)=\vert\{\outcome{\driverset}{\singleplan{}}\mid\singleplan{}\in\scenarioset\}\vert$ if $\scenarioset$ is differentiated, and $\diversityindicator(\scenarioset)=0$ otherwise).
The problem maximises $\diversityindicator$, and a set maximises it exactly when it is a maximal differentiated set, because such a set holds one scenario for each outcome that can be reached within the budget.
In Shell's exercise with the budget $\costthreshold=3$, the solution is $\{\singleplan{c},\singleplan{d}\}$, the continuity and the disruption narratives, which together are Shell's scenario set.

To ground the criteria in PDDL3, \Cref{lst:shell-criteria} collects Shell's constraints for the exercise in a PDDL3 constraints block and appends it to the problem in \Cref{lst:shell-pddl}.
Three of the four criteria that judge one narrative have a PDDL3 form on Shell: internal consistency is the invariant, and relevance and challenge are \texttt{at end} constraints, since $\ltlalways\ltleventually\psi$ holds exactly when $\psi$ holds at the final position.
Plausibility's conditions read actions, as do the action-trajectory constraints of \citet{bonassi2022action}.
Therefore, the block states them with action atoms.
For conditions (1) to (3) need to express them, since every actor's action raises its actor's objective, reads a fluent set earlier and leaves no state to recur.
While condition (4) requires to be expressed by asking that a later action of an actor reads every happening, and the only action that reads the war is the resupply, while the only one that reads the cartel's split is the growth, so each happening gets the one constraint that names its reader.

\begin{lstlisting}[morecomment={[l]{;}},caption={Shell's criteria as PDDL3 constraints, to append to the problem of \Cref{lst:shell-pddl}.},label={lst:shell-criteria}]
(:constraints (and
  ; internal consistency: an embargo in force requires a coordinated cartel,
  ; the invariant of the subject
  (always (imply (embargoed) (coordinated)))
  ; plausibility: every happening is read by a later action of an actor,
  ; the outbreak by the resupply and the split by the growth
  (sometime-after (outbreak ally neighbour) (resupply importer ally))
  (sometime-after (split) (growth importer))
  ; relevance: the concern, the importer's capacity, ends away from its
  ; initial value
  (at end (not (= (capacity) 1)))
  ; challenge: the drivers do not both end at the conventional outcome,
  ; a coordinated cartel at peace
  (at end (or (not (coordinated)) (war)))))
\end{lstlisting}

Differentiation compares narratives with one another.
A planner iteratively updates constraints by forbidding the generated outcome.
\Cref{lst:shell-forbid} shows the iteration on Shell at budget $\costthreshold=3$: the first call returns the disruption narrative, with the outcome $\langle\mathit{true},\mathit{true}\rangle$, which the first appended line forbids; the second call returns the continuity narrative, with the outcome $\langle\mathit{false},\mathit{false}\rangle$, which the second line forbids; and the third call finds no narrative within the budget, so the set is complete with two members.
The one remaining outcome, $\langle\mathit{false},\mathit{true}\rangle$, is out of reach at this budget. 

\begin{lstlisting}[morecomment={[l]{;}},caption={Differentiation by iteration on Shell at budget $3$: the seed is the challenge line of \Cref{lst:shell-criteria}, and each call appends the line that forbids the outcome it returned.},label={lst:shell-forbid}]
; call 1 returns the disruption narrative, outcome <true, true>: forbid it
  (at end (or (not (coordinated)) (not (war))))
; call 2 returns the continuity narrative, outcome <false, false>: forbid it
  (at end (or (coordinated) (war)))
; call 3 returns no narrative within the budget: the set is complete
\end{lstlisting}

\subsection{Planning as Satisfiability}\label{sec:engine:sat}

Planning as satisfiability~\cite{kautz1992planning} solves a planning problem by issuing a series of SAT queries $\satformula{0},\satformula{1},\satformula{2},\ldots$, where each $\satformula{n}$ checks whether a plan of length $n$ exists.
The encoding introduces a separate copy of every fluent at every time step, so the state $\state{i}$ is given by the truth values of these copies at step $i$.
It introduces a Boolean variable for each action at each step, indicating whether that action is executed at that time.
The query is $\satformula{n}\mathrel{:=}\satinit(\state{0})\wedge\bigwedge_{i=0}^{n-1}\sattrans(\state{i},\state{i+1})\wedge\satgoal(\state{n})$.
Here, $\satinit$ specifies the initial state and $\satgoal$ specifies the goal.
The component $\sattrans$ represents the transition function $\transitionfn$.
Any model of $\satformula{n}$ (i.e., any satisfying assignment) corresponds to a plan, obtained by reading off which actions are true at each step.
The solver considers $n=0,1,2,\ldots$ until it finds a satisfiable query, so the first plan returned is of minimal length.
In our setting, the trajectory constraints $\trajconstraints$ of \Cref{sec:engine:criteria} replace the goal, and the budget $\costthreshold$ bounds a narrative's cost.
Every constraint $\theta\in\trajconstraints$ is evaluated over the trajectory $\langle\state{0},\ldots,\state{n}\rangle$, and the last conjunct of the query is the budget constraint, which sums the costs of the actions that occur:

\begin{equation*}
 \satformula{n}\mathrel{:=}\satinit(\state{0})\wedge\bigwedge_{i=0}^{n-1}\sattrans(\state{i},\state{i+1})\wedge\bigwedge_{\theta\in\trajconstraints}\theta(\state{0},\ldots,\state{n})\wedge\Bigg(\Big(\sum_{i=1}^{n}\sum_{\action{}\in\allaction}\costfn(\action{})\cdot\mathrm{occ}_i(\action{})\Big)\leq\costthreshold\Bigg),
\end{equation*}

An action atom $\action{}$ at step $i$ is the variable $\mathrm{occ}_i(\action{})$, and a state atom $\state{}[\fluent]=\avalue$ at step $i$ is the copy of $\fluent$ at that step, where $\mathrm{occ}_i(\action{})$ is $1$ if the action $\action{}$ occurs at step $i$ and $0$ otherwise.
A trajectory constraint becomes a formula over these variables by unrolling its temporal operators over the steps $0,\ldots,n$.
A step may hold no action, provided that every later step holds none either (i.e., $\bigvee_{\action{}\in\allaction}\mathrm{occ}_i(\action{})\rightarrow\bigvee_{\action{}\in\allaction}\mathrm{occ}_{i-1}(\action{})$), so each model of $\satformula{n}$ corresponds to a scenario with at most $n$ actions and a cost within the budget.
We issue a single query with $n$ large enough to fit every in-budget scenario in $n$ steps, and the budget constraint bounds the cost.
Enumerating the models of this query yields all in-budget scenarios.

The planner proposed by \citet{abdelwahed2026behaviour} can solve this scenario-planning problem.
Because $\fbixoperator{}$ uses a feature model as formulas, a set of plans is diverse when the plans differ on the features~\cite{abdelwahed2026behaviour}.
It appends formulas to a planning-as-satisfiability query, so it can take our constraints directly, and a narrative's outcome can serve as its behaviour.
$\fbixoperator{}$ finds a plan, forbids that plan's behaviour, and calls the solver again, so it needs one call per outcome.
We use ASP instead, because it gives the same result in one call.
ASP is a declarative framework in which a problem is written as logic rules, and a solver returns answer sets (i.e., models of the rules, each of which is a solution).
It first grounds the rules with the problem's constants, and then solves the ground program.
An ASP solver like clingo~\cite{gebser2019multi} enumerates and counts all answer sets of a program, which we use to generate all scenarios within the budget.
With projection, it returns one answer set for each distinct assignment over a chosen subset of atoms~\cite{gebser2009solution}.
Projecting on the outcome returns one scenario per outcome, as $\fbixoperator{}$ would, without looping solver calls.
The next subsection gives the implementation.

\subsection{The ASP Implementation}\label{sec:engine:asp}
An ASP program is a set of rules, and the solver's task is to find its \emph{answer sets}, sets of atoms the rules support.
A rule has a head and a body, written $\mathtt{head\ \mbox{:-}\ body}$, and it reads: the head holds whenever the body holds.
The body is a list of atoms, each possibly under $\mathtt{not}$ (i.e., the atom is not derived), and the capitalised terms are variables that the solver instantiates with every object before solving (e.g., $\mathtt{holds}(V,X,t)$ stands for one atom per fluent $V$ and value $X$).
Two rule forms have special names: a \emph{fact} has no body, so its head always holds, and an \emph{integrity constraint} has no head, so its body may never hold, removing any answer set in which it does.

We adopt the encoding of the planning problem of plasp~\cite{dimopoulos2019plasp}, which converts a PDDL planning problem into ASP facts and solves it with an ASP planning encoding\footnote{The implementation of plasp that we use is available at \url{https://github.com/MFaisalZaki/ASPPlanners}.}, and we append to it one program per criterion of \Cref{sec:engine:criteria}.
The analyst writes the planning problem $\planningtask$ for a topic in PDDL3, which the plasp translates into the $\satinit$ and $\sattrans$ encoding.
For the scenario criteria, each is implemented in a logic program, and the programs are then appended to construct $\satformula{n}$.
Each program realises its constraint as one or more \emph{laws}, named patterns that no scenario may contain.
A law $\ell$ is defined by the fact $\mathtt{law}(\ell)$ and includes rules that derive $\mathtt{broken}(\ell,t)$ whenever the pattern that violates the constraint appears at step $t$. 
All laws share one integrity constraint: the solver cannot produce a narrative that breaks any law. 
This means the solver eliminates such narratives during search rather than filtering them out later.
Differentiation is a projection directive, which we cover later in this section.
We will first explain the planning process, then outline the rules for each criterion in the order given in \Cref{sec:engine:criteria}.

\paragraph{\textbf{The planning encoding.}}
plasp encodes $\satformula{n}$ as ASP rules as follows (\Cref{lst:encoding}).
The atom $\mathtt{holds}(v,x,t)$ represents the copy of fluent $v$ at step $t$ with value $x$, and $\mathtt{occurs}(a,t)$ denotes the variable for action $a$ at step $t$.
The program is divided into three sections~\cite{gebser2019multi}.
The \texttt{base} section captures $\satinit$, and the \texttt{step(t)} section captures $\sattrans$ from step $t-1$ to $t$.
The laws are split between \texttt{step(t)} and \texttt{check(t)}.
A law whose pattern is visible at its own step detects it in \texttt{step(t)}, and a law that needs the whole narrative (the causal chains of the intentional law, relevance and challenge) detects it in \texttt{check(t)}.
The \texttt{check(t)} section holds the one constraint that enforces every law, grounded once at the last step, identified by the external atom $\mathtt{query}(t)$.
For clarity, we use $v$, $x$, and $a$ instead of plasp's terms $\mathtt{variable}(v)$, $\mathtt{value}(\mathtt{variable}(v),x)$, and $\mathtt{action}(a)$.
The plasp implementation omits the initial value of a Boolean fluent that starts false unless an action reads it as false.
The second rule of the \texttt{base} section ensures that every fluent without an initial value is false at step $0$, as subsequent rules read these false values from the state.
In the \texttt{step(t)} section, the choice rule ensures that at most one action occurs at each step.
The subsequent integrity constraint (i.e., a headless rule that eliminates any answer set whose body holds) enforces the action's precondition.
The $\mathtt{caused}/3$ rules implement the effects, and the following rule serves as the frame axiom.
The encoding has exactly $\costthreshold$ steps, so a narrative with fewer than $\costthreshold$ actions leaves its last steps empty (i.e., with no action).
The $\mathtt{started}/1$ and $\mathtt{ended}/1$ rules guarantee that a narrative begins at the first step and ends at its first empty step, so each such plan is generated exactly once.
The last two rules of the \texttt{step(t)} section reject an actor's action that changes no fluent, so a narrative never repeats an action to no effect.
The final rule of \Cref{lst:encoding} is common to all laws and eliminates any answer set that breaks a law.
In that rule, the atom $\mathtt{waived}(L)$ lets an analyst disable any law to suit their needs (i.e., adding the fact $\mathtt{waived}(\ell)$ stops the law $\ell$ from removing any answer set), and the narratives the engine then accepts meet only the criteria whose laws remain.

\begin{lstlisting}[caption={The ASP encoding for the planning problem (partial; the full encoding is in the supplementary code).},label={lst:encoding}]
#program base.
% I: the initial state, where a fluent with no initial value is false
holds(V, X, 0) :- initialState(V, X).
holds(V, false, 0) :- variable(V), not initialState(V, _).
#program step(t).
% T: at most one action, its precondition, its effects and the frame axiom
{ occurs(A, t) : action(A) } 1.
:- occurs(A, t), precondition(A, V, X), not holds(V, X, t-1).
caused(V, X, t) :- occurs(A, t), postcondition(A, _, V, X).
modified(V, t) :- caused(V, _, t).
holds(V, X, t) :- caused(V, X, t).
holds(V, X, t) :- holds(V, X, t-1), not modified(V, t).
% a narrative starts at the first step and ends at its first empty step
acted(t) :- occurs(_, t).
started(t) :- acted(t).
started(t) :- started(t-1).
ended(t) :- started(t-1), not acted(t).
ended(t) :- ended(t-1).
:- acted(t), ended(t-1).
:- acted(t), t > 1, not started(t-1).
% an action of an actor must change a value it writes
moved(t) :- caused(V, X, t), not holds(V, X, t-1).
:- occurs(A, t), not happening(A), not moved(t).
#program check(t). #external query(t).
:- broken(L, _), not waived(L). % L: no law may be broken
\end{lstlisting}

\paragraph{\textbf{Internal consistency.}}
The engine supports invariants of the form $\psi=\neg(\fluent_1=x_1\wedge\fluent_2=x_2)$, indicating that two values cannot hold simultaneously, which the topic lists as facts $\mathtt{incompatible}(\fluent_1,x_1,\fluent_2,x_2)$.
This pairwise form is an implementation limit, which accepts any logical formula over $\allfluents$.
The \emph{repeat} law rejects a narrative in which two actors' actions share a schema, actor and target, unless the topic lists the schema as one that may recur, as a fact $\mathtt{sustainedClass}(N)$, so the domain needs no such fluent and lists only the schemas that recur.
\Cref{lst:consistency} shows the two laws: the atom $\mathtt{toldAs}(N,P,t)$ holds when the action at step $t$ is an actor's, has the schema $N$ and pairs its actor with its target in $P$, and $\mathtt{toldBefore}(N,P,t-1)$ holds when an earlier step has the same schema, actor and target.
In Shell's exercise, the topic lists $\mathtt{incompatible}(\mathit{embargoed},\mathit{true},\mathit{coordinated},\mathit{false})$.
In the narrative $\langle\mathit{outbreak},\mathit{resupply},\mathit{embargo},\mathit{split}\rangle$, the rule identifies $\mathtt{holds}(\mathit{embargoed},\mathit{true},4)$ and $\mathtt{holds}(\mathit{coordinated},\mathit{false},4)$, derives $\mathtt{broken}(\mathit{incompatible},4)$, and the narrative is rejected.

\begin{lstlisting}[caption={Internal consistency: incompatible values and recurring signatures.},label={lst:consistency}]
#program base.
law(incompatible). law(repeat).
broken(incompatible, 0) :- incompatible(V1, X1, V2, X2), holds(V1, X1, 0), holds(V2, X2, 0).
#program step(t).
broken(incompatible, t) :- incompatible(V1, X1, V2, X2), holds(V1, X1, t), holds(V2, X2, t).
broken(repeat, t) :- toldAs(N, P, t), toldBefore(N, P, t-1), not sustainedClass(N).
\end{lstlisting}

\paragraph{\textbf{Relevance.}}
The topic lists the concern fluents as facts $\mathtt{concern}(\fluent)$.
A concern fluent changes when its step-$0$ value no longer holds at the last step; \Cref{lst:relevance} implements the constraint $\critformula{rel}$, and its law fires only when the topic lists at least one concern.
In Shell's exercise, the topic lists the fact $\mathtt{concern}(\mathit{capacity})$.
In the disruption narrative, $\mathtt{holds}(\mathit{capacity},\mathit{normal},0)$ holds, but $\mathtt{holds}(\mathit{capacity},\mathit{normal},3)$ does not, so the rules derive $\mathtt{changed}(3)$ and the law is not broken.
In $\singleplan{r}$, the capacity remains at $\mathit{normal}$ at step $3$, so $\mathtt{changed}(3)$ is not derived, and the rules derive $\mathtt{broken}(\mathit{relevant},3)$ instead.

\begin{lstlisting}[caption={Relevance.},label={lst:relevance}]
#program base.
law(relevant).
#program check(t).
changed(t) :- query(t), concern(V), holds(V, X, 0), not holds(V, X, t).
broken(relevant, t) :- query(t), concern(_), not changed(t).
\end{lstlisting}

\paragraph{\textbf{Challenge.}}
The topic lists $\expectedoutcome$ as facts $\mathtt{conventional}(\fluent,x)$, and the law of \Cref{lst:challenge} fires only when the topic lists them.
The rule's body uses a conditional literal (i.e., a literal that must hold for every instance of the condition after the colon), so the law is broken exactly when every driver ends at its conventional value.
In Shell's exercise, the topic lists $\mathtt{conventional}(\mathit{coordinated},\mathit{true})$ and $\mathtt{conventional}(\mathit{war},\mathit{false})$.
For the disruption narrative, $\mathtt{holds}(\mathit{coordinated},\mathit{true},3)$ is true, but $\mathtt{holds}(\mathit{war},\mathit{false},3)$ is not, so the body fails and $\mathtt{broken}(\mathit{challenging},3)$ is not derived.

\begin{lstlisting}[caption={Challenge.},label={lst:challenge}]
#program base.
law(challenging).
#program check(t).
broken(challenging, t) :- query(t), conventional(_, _), holds(V, X, t) : conventional(V, X).
\end{lstlisting}

\paragraph{\textbf{Plausibility.}}
Each condition of \Cref{def:plausible} is a law of the same name, which are shown in \Cref{lst:plausible}.
The engine reads an action's actor and target from its first two arguments, and the topic names the types of objects that are actors, which gives the facts $\mathtt{agent}(\anactor)$, and lists each happening as a fact $\mathtt{happening}(\action{})$.
The topic lists each actor's preferences as facts $\mathtt{preference}(\anactor,\fluent,\avalue,\aweight)$, one per preference of $\objectives$.
The term $\mathtt{gainTerm}(A,V,W,t)$ is used when an action at step $t$ moves the objective of $A$ by $W$ through a fluent $V$ that it wrote. 
If the action changes $V$, $\mathtt{beatFlip}(V,t)$ holds. 
If it changes a numeric value of $V$ by $D$, then $\mathtt{beatDelta}(V,D,t)$ holds. 
The first two rules score a Boolean preference for whether the action makes it true or false. 
The third rule gives a score times the change in a numeric value, so the results of a step add up to $\objectiveof{A}(\state{t})-\objectiveof{A}(\state{t-1})$, the change of the objective of $A$ at that step. 
The term $\mathtt{raises}(A,t)$ shows how $\mathrm{raises}(\action{t},A)$ is realised, requiring the total to be positive.
The atom $\mathtt{supports}(t,t')$ is a causal link from step $t$ to step $t'$, and $\mathtt{answers}(t',t)$ holds when the action at $t$ replies to the one at $t'$.
The rules for $\mathtt{servesActor}(A,t)$ focus on how actions can help an actor, $A$, at time $t$. An action helps $A$ if it improves $A$'s goals, responds to something $A$ did ($\mathtt{motivated}(t)$), or another actor's action uses what they wrote to assist $A$.
An actor's action is intentional if it helps the actor or responds to something. The atom $\mathtt{explained}(t,R)$ tracks this action and its reason $R$. A topic can also create $\mathtt{explained}$ atoms with its own reasons, which also count as $\mathtt{motivated}$.
The chain rules are evaluated at the end in the \texttt{check(t)} section.

\begin{lstlisting}[caption={The intentional and connected laws (partial; the full encoding, with the loop-free law, is in the supplementary code).},label={lst:plausible}]
#program base.
law(intentional). law(connected). law(loop_free).
#program step(t).
gainTerm(A, V, W, t) :- preference(A, V, X, W), X != up, X != down,
 beatFlip(V, t), holds(V, X, t).
gainTerm(A, V, -W, t) :- preference(A, V, X, W), X != up, X != down,
 beatFlip(V, t), holds(V, X, t-1).
gainTerm(A, V, W*D, t) :- preference(A, V, up, W), beatDelta(V, D, t).
raises(A, t) :- gainTerm(A, _, _, t), #sum{ W,V : gainTerm(A, V, W, t) } > 0.
explained(t, reply(TT)) :- agentActed(t), answers(TT, t).
broken(connected, t) :- agentActed(t), beatStarted(t-1), not linked(t).
#program check(t).
motivated(T) :- explained(T, reply(_)).
servesActor(A, T) :- query(t), raises(A, T).
servesActor(A, T) :- query(t), agentActed(T), actor(A, T), motivated(T).
servesActor(A, T1) :- query(t), supports(T1, T2), agentActed(T2), servesActor(A, T2).
explained(T, raises) :- query(t), agentActed(T), actor(A, T), raises(A, T).
explained(T, chain(T2)) :- query(t), agentActed(T), actor(A, T), supports(T, T2),
 agentActed(T2), servesActor(A, T2).
intentional(T) :- explained(T, _).
broken(intentional, T) :- query(t), agentActed(T), not intentional(T).
happeningRead(TT) :- query(t), supports(TT, T2), agentActed(T2).
broken(connected, TT) :- query(t), occurs(A, TT), happening(A), not happeningRead(TT).
\end{lstlisting}

In Shell's exercise, the topic lists $\mathtt{agent}(\mathit{cartel})$, $\mathtt{agent}(\mathit{importer})$, $\mathtt{agent}(\mathit{ally})$, $\mathtt{happening}(\mathit{outbreak})$, $\mathtt{happening}(\mathit{split})$, and the facts $\mathtt{preference}(\mathit{importer},\mathit{resupplied},\mathit{true},1)$, $\mathtt{preference}(\mathit{importer},\mathit{capacity},\mathit{up},1)$ and $\mathtt{preference}(\mathit{cartel},\mathit{embargoed},\mathit{true},1)$\footnote{The supplementary code represents $\mathit{capacity}$ as a number, so the growth raises the importer's objective by raising the capacity by one}. 
The intentional law holds for every narrative the planning encoding produces, because every action by the importer and the cartel raises its actor's objective.
The connected law is the only plausibility law that rejects a narrative at $\costthreshold\leq4$.
At $\costthreshold=3$, the planning encoding produces 14 narratives, counting the empty one.
The connected law is broken in ten of them, each with an outbreak or a split from which no causal link leads to a later action, because only the resupply reads the war and only the growth reads the split.
In three of the ten, it is the only broken law, namely $\langle\mathit{outbreak},\mathit{split},\mathit{growth}\rangle$, $\langle\mathit{split},\mathit{outbreak},\mathit{growth}\rangle$ and $\langle\mathit{split},\mathit{growth},\mathit{outbreak}\rangle$; the other seven are also irrelevant, since their capacity stays normal.
In the disruption narrative, the rules derive $\mathtt{explained}(2,\mathtt{raises})$ and $\mathtt{explained}(3,\mathtt{raises})$, so no plausibility law is broken.

\paragraph{\textbf{Differentiation and scenario generation.}}\label{sec:engine:generation}
The engine uses projected enumeration to generate scenarios with different outcomes.
With projection, clingo returns one answer set for each assignment to a chosen set of atoms~\cite{gebser2009solution}.
The topic lists the drivers as facts of the form $\mathtt{driver}(\fluent)$.
\Cref{lst:differentiation} derives the atom $\mathtt{outcome}(\fluent,x)$ for a driver $\fluent$ with value $x$ at the final step.
The \texttt{\#project} directive makes the solver return one answer set per distinct set of outcome atoms (i.e., per distinct combination of driver values, since each driver takes exactly one value in the final state, so the set of outcome atoms is the outcome $\outcome{\driverset}{\singleplan{}}$).
The engine's program is the planning encoding of \Cref{lst:encoding} together with the laws of \Cref{lst:plausible,lst:consistency,lst:relevance,lst:challenge} and this listing, so its answer sets are the scenarios within the budget.
This program is the same for every topic.
Projection never returns a second answer set with an outcome it has already returned, much as $\fbixoperator{}$ forbids a behaviour once it has a plan for it.
The difference is that clingo does this within one call rather than one call per outcome.

\begin{lstlisting}[caption={Differentiation by projection on the drivers.},label={lst:differentiation}]
#program check(t).
outcome(V, X) :- query(t), driver(V), holds(V, X, t).
#project outcome/2.
\end{lstlisting}

In Shell's exercise with the budget $\costthreshold=3$, clingo returns two answer sets, the continuity narrative $\singleplan{c}$ and the disruption narrative $\singleplan{d}$, as \Cref{sec:engine:differentiation} anticipated.
In \Cref{sec:case-study}, we apply the engine to geopolitical conflicts and to supply chains.

%% file: sections-v5/04-case-studies.tex
\expandafter\def\expandafter\UrlBreaks\expandafter{\UrlBreaks\do\-}
\section{Case Studies}\label{sec:case-study}

In \Cref{sec:engine}, we introduced the engine and applied it to Shell's exercise.
Here, we present a step-by-step guide showing how an analyst can reuse the engine for a new topic, and we illustrate the procedure for 2 topics from different domains.
The procedure has 4 steps, and each writes a part of the topic $\topic=\langle\planningtask,\actors,\objectives,\concernset,\driverset,\expectedoutcome\rangle$: the analyst develops the planning domain (Step 1) and poses the question that turns it into a topic (Step 2), a script grounds the initial state (Step 3), and the engine generates the set (Step 4).
Each topic is then covered in its own subsection, with 1 paragraph per step.

\paragraph{\textbf{Step 1: develop the planning domain.}}
The analyst writes the planning domain, denoted $\planningtask$, in PDDL (Planning Domain Definition Language). 
This includes specifying its objects, its fluents (Boolean or numeric), and action schemas that outline preconditions and effects. 
The first parameter of a schema names its actor. 
When the action targets an object, the second names it.
A happening names no actor.
The domain describes what can occur within the subject matter rather than detailing what the organisation inquires about it.
Therefore, 1 domain can address all questions related to the same subject. 
When the subject has an event taxonomy, each schema can represent a specific event type. 
If no taxonomy exists, the analyst compiles schemas from existing literature. 
The following 2 subsections provide 1 example of each approach. 
This step is conducted manually, and the engine evaluates narratives based solely on the topic. 
Consequently, scenarios cannot include events outside the domain.

\paragraph{\textbf{Step 2: pose the question and write the topic.}}
Scenario planning begins with what an organisation needs to know.
\citet{schoemaker1995scenario} starts by defining the analysis scope: name the organisation and the decision it faces, then state its concerns (what must hold at the end of a scenario) and its drivers (uncertainties whose outcomes distinguish scenarios).
The question turns the domain into a topic.
It decides which domain fluents are concerns $\concernset$ and which are drivers $\driverset$, and it names the organisation from whose side the criteria read a narrative.
It also sets the actors' objectives $\objectives$ (i.e., their weighted preferences) and the topic's invariants (trajectory constraints in $\trajconstraints$).
Every element in the topic translates directly into facts and rules, except $\planningtask$, which translates into plasp rules.
In Shell's exercise, the concern was the importer's ability to buy oil, and the drivers were cartel coordination and regional war.
After this step, the record includes every element of $\topic$ except the initial state of $\planningtask$ and the conventional outcome $\expectedoutcome$, which the record supplies.

\paragraph{\textbf{Step 3: ground the initial state.}}
A script compiles a \emph{board} (i.e., the planning problem for the topic, with an initial state read from the record, a list of objects, and a date).
The drivers' values on the board yield the conventional outcome $\expectedoutcome$, completing the topic.
1 script serves any combination of objects and dates, although a board is only as accurate as the thresholds the analyst sets for reading the record.

\paragraph{\textbf{Step 4: generate the scenario set.}}
The analyst chooses a budget, and the engine solves the scenario planning problem with clingo, projecting on the outcome atoms (\Cref{sec:engine:generation}).
These 4 steps make up the entire method.
For each case study, we include a retrospective interpretation of the scenario set (that is, we anchor the board to a past date and verify which outcomes from the set actually occurred afterwards).
This way, we use it to illustrate what the set would have produced if the engine had run on the board's date, so we present it alongside the scenario set rather than as a standalone test.

We apply the same procedure to 2 topics, with each drawing its planning domain from a distinct source.
\Cref{tab:topics} lists all elements of the 2 topics, and the 2 subsections explain where each comes from.
The geopolitical-conflict domain (\Cref{sec:case-study:conflict}) draws on the CAMEO event taxonomy~\cite{gerner2002cameo}, while the supply-chain domain (\Cref{sec:case-study:supply-chain}) adopts its action schemas from the supply-chain risk literature.
We process every topic the same way.
Its board date is 1 January 2021, and the script reads sources through the previous day.
We use a 5-year window and a budget of $5$ per topic.
We chose $5$ because it is the smallest budget at which an outcome can hold 2 wars: a war needs a mobilisation and a strike after a threat (\Cref{sec:case-study:conflict}), so a budget of $4$ reaches the same 5 single-war outcomes as a budget of $5$ and no outcome with 2 wars.
\Cref{tab:runs} lists each run's size and cost, together with the number of outcomes at budgets $4$, $6$ and $7$.
The tables in this section are from single-thread runs, which always return the same set.
With multiple threads, the solver returns the same outcomes but may select a different narrative for an outcome, since projection returns the first narrative it reaches.
Using dated news reports and announcements, we check whether each outcome occurred by 1 January 2026.

\begin{table}[t]
\centering
\footnotesize
\renewcommand{\arraystretch}{1.15}
\begin{tabular}{@{}>{\raggedright\arraybackslash}p{0.8cm}>{\raggedright\arraybackslash}p{2.5cm}>{\raggedright\arraybackslash}p{5.05cm}>{\raggedright\arraybackslash}p{5.05cm}@{}}
\toprule
Step & Element of $\topic$ & Geopolitical conflicts (\Cref{sec:case-study:conflict}) & Supply chains (\Cref{sec:case-study:supply-chain}) \\
\midrule
1 & $\planningtask$: domain & 65 action schemas over 6 countries, 1 per CAMEO event type; fluents for the pair ladder, 9 ties, discontent and the economy. & 17 action schemas over the lines, the maker, the rivals, the supplier and the authorities, 6 of them happenings; fluents for closure, attack, diversion, shortage, orders and stops, and 4 numeric levels; flying parts in, switching source and escorting ships require a fact the analyst states on the board (air-transportable parts, a qualified second source, spare warships), false on every board here. \\
 & actors $\actors$ & the 6 countries; every action is a country's, and its target is the country it is directed at & the lines, the maker, the rivals, the supplier and the authorities; the 6 happenings have no actor \\
\midrule
2 & $\concernset$ & the partnership of each pair and the economy of each country & the maker's stock \\
 & $\driverset$ & whether each of the 15 pairs is at war & whether the route is closed, whether the lines have diverted, and whether the maker has stopped a line \\
 & objectives $\objectives$ & the week's motives and the maintained table; 1 preference per motive, weighted by the motive's weight & lines want ships moving ($2$) and rates up ($1$ per level); buyers want no missed delivery ($3$), their lines running ($2$) and stock up ($1$ per unit); the supplier wants fillable orders ($1$); the authorities want routes open ($2$) and safe ($2$) \\
 & $\trajconstraints$ & escalation, stranger, reversal and capitulation & none; the invariants are in the preconditions \\
\midrule
3 & $\planningtask$: initial state & the board of 1 January 2021, read from GDELT, the maintained tables and the economic figures & 1 board per route of 1 January 2021, read from PortWatch, the pressure index and GDELT \\
 & $\expectedoutcome$ & no pair at war (the board's values) & the route open, no diversion, no stop (the board's values) \\
\bottomrule
\end{tabular}
\caption{The 2 topics of this section.}
\label{tab:topics}
\end{table}

\begin{table}[t]
\centering
\footnotesize
\begin{tabular}{@{}llrrrrrr@{}}
\toprule
 & & \multicolumn{4}{c}{Outcomes at budget} & \multicolumn{2}{c}{Budget $5$} \\
Topic & Board (1 January 2021) & $4$ & $5$ & $6$ & $7$ & Time & Memory \\
\midrule
Conflicts & 6 countries & $5$ & $8$ & $9$ & $11$ & $8.4$ s & $350$ GB \\
Supply chain & Suez Canal & $3$ & $3$ & $3$ & $3$ & $0.07$ s & $36$ MB \\
 & Panama Canal & $3$ & $3$ & $3$ & $3$ & $0.07$ s & $36$ MB \\
 & Baltimore & $2$ & $3$ & $3$ & $3$ & $0.07$ s & $37$ MB \\
 & Strait of Hormuz & $3$ & $3$ & $3$ & $3$ & $0.07$ s & $36$ MB \\
\bottomrule
\end{tabular}
\caption{The number of outcomes at budgets $4$, $5$, $6$ and $7$, and the clingo runtime and peak memory at budget $5$, as means of 10 single-thread runs on an Apple M1 laptop (8 cores, 16 GB, clingo 5.8.2).}
\label{tab:runs}
\end{table}

\subsection{Geopolitical Conflicts}\label{sec:case-study:conflict}

A geopolitical conflict is a dispute between countries that can escalate from verbal disagreements to sanctions and even military action, potentially involving nations that were not originally part of the conflict.
This analysis focuses on 6 countries: China, Iran, Israel, Russia, Ukraine, and the United States.
We examine events among these nations from 1 January 2021 to 1 January 2026.
Our analysis uses the CAMEO (Conflict and Mediation Event Observations) framework~\cite{gerner2002cameo}, which categorises state interactions as described in news reports.
Each action schema reflects a specific event type within this taxonomy.

\paragraph{\textbf{The planning domain (Step 1).}}
The planning domain includes 6 countries and 65 action schemas.
Each schema is tagged with the CAMEO root it abstracts (one of 20 top-level event types), and every root has at least 1 schema, spanning public statements (01), consulting (04), threatening (13), coercing (17), and fighting (19).
Fluents place each country pair on a 4-rung ladder ($\mathit{neutral}$, $\mathit{tense}$, $\mathit{hostile}$, $\mathit{in\text{-}conflict}$), record ties (partnership, alliance, talks, border, dependence, aid, sanctions, massed forces, occupation), track population discontent, and assign each country a numeric $\mathit{economy}$.
Most actions shift a pair 1 rung or change a tie: threats raise tense to hostile; strikes/attacks raise hostile to conflict; ceasefires reverse this; de-escalation requires talks.
Threats depend on economies, and each country chooses its own events.
The preconditions explain what makes an action believable. 
For example, an action with a high intensity of 4 needs either 2 sides that are already at war or forces focused on the target. 
A covert operation requires either a hostile pair or a pair that is at war. 
Forming an alliance or a new partnership needs states that are either in discussions or already partners.

\paragraph{\textbf{The question and topic (Step 2).}}
An analyst poses a question for an organisation trading across 6 countries: which wars between them are possible, and how would each affect trade?
It focuses on 15 trade partnerships and 6 economies.
A scenario matters only if a partnership forms or breaks, or if an economy ends differently than it began.
The drivers are whether each of the 15 possible country pairs is at war.
This setup treats a settlement and a frozen stalemate as the same outcome and includes 1 scenario for every war combination within the budget, so the set grows quickly with the number of pairs.
The concerns are $\concernset=\{\mathit{partners},\mathit{economy}\}$ for each pair and country, and the drivers $\driverset$ are $\mathit{in\text{-}conflict}(s,t)$ for each country pair $s,t$.
Countries are the actors; the first argument of each schema names the acting country, and the second the country it is directed at.
Their objectives come from motives read from the record and a maintained table (i.e., each a preference with the motive's weight).
Each schema has an \emph{intensity} $1$--$4$: $4$ war, $3$ skirmish/covert operation, $2$ sanctions/threats/demands/mobilisation/arming, and $1$ renouncing an alliance or a diversionary dispute.
Schemas are \emph{coercive} from intensity $2$.
A \emph{tie} holds between 2 countries when any standing relation links them.
This topic adds its own laws.
The \emph{escalation} law allows the highest-intensity action only after earlier coercion between the same pair (either direction).
The \emph{stranger} law allows coercion only between countries that share a tie or have interacted before.
The \emph{reversal} law forbids a country from undoing the pair's last shift if the other has not directed any action at it since.
In \Cref{lst:conflict-laws}, $\mathtt{directed}(C,S,T,t)$ means the step-$t$ action has schema $C$, actor $S$, and target $T$, and $\mathtt{coerciveBeat}(S,T,t)$ means that action is coercive.
Without intensities and ties, these laws cannot fire, so the engine would allow a sudden attack between unrelated countries with no prior pressure.
The \emph{capitulation} law permits a concession (that is, granting a demand, giving in to pressure, or yielding under duress) only if the conceding state has previously been pressured by the other party (i.e., the other coerced it at intensity $2$ or higher earlier in the narrative, imposed sanctions on it, had forces massed against it, or was hostile toward it or at war with it at the outset).

\begin{lstlisting}[caption={The escalation and stranger laws (partial; the full encoding is in the supplementary code).},label={lst:conflict-laws}]
#program base.
law(escalation). law(stranger).
#program step(t).
broken(escalation, t) :- directed(C, S, T, t), actionIntensity(C, M),
 maxIntensity(M), not pairCoerced(S, T, t-1).
broken(stranger, t) :- coerciveBeat(S, T, t), not tie(S, T, t-1), not interacted(S, T, t-1).
\end{lstlisting}

\paragraph{\textbf{The initial state (Step 3).}}
The script uses GDELT (Global Data on Events, Location, and Tone), which codes global news into daily CAMEO events~\cite{leetaru2013gdelt}.
Each event includes an actor, target, root, and \emph{Goldstein tone} (a $-10$ to $+10$ conflict--cooperation score).
It keeps events among 6 countries in the 7 days before the date.
A pair is in conflict if both sides use force against the other (CAMEO root 19 or 20) at least $280$ times that week.
Otherwise, the week's mean tone classifies the pair as hostile ($\le -0.5$), tense ($\le 1.25$), or \emph{cooperative} ($\ge 1.40$).
A cooperative pair is ladder-neutral and recorded as trade partners; a pair is also neutral if its tone is between $1.25$ and $1.40$ or its week has too few events.
A tense or hostile pair has talks when consultations are at least $17\%$ of events.
We calibrated cutoffs on pre-date weeks in 2020 using 8 pairs with clear relations, placing each threshold midway between the closest pairs on either side.
For instance, Armenia--Azerbaijan (October 2020) and Iran--U.S. (January 2020) were at war; China--U.S. were not, despite $38$ and $39$ force events, reflecting noisy machine coding.
1 calibration case misclassifies India--China as neutral because their week was dominated by talks.
Maintained tables supplement what news misses (borders, alliances, dependence, sanctions, strike reach, revisionist states), with entries dated by era.
Economic inputs use the most recent pre-dated annual figures (for 2021, 2019), and motives draw on ongoing wars, sanctions regimes, and that week's economic coverage.
On 1 January 2021, no pair is in conflict.
Iran is hostile to Israel, Ukraine, and the United States, and is in talks with the United States.
Russia is tense and in talks with Israel, Ukraine, and the United States; China and the United States are also tense and in talks.
All other pairs are neutral.
Israel is allied with the United States.
China trades with the other 5; Iran trades with Russia; and Israel, Ukraine, and the United States trade with each other.
The United States sanctions Iran and Russia, and Russia and Ukraine sanction each other.
Populations in Iran, Israel, Ukraine, and the United States are discontented.
The drivers indicate no war, which is the conventional outcome.

\begin{table}[h]
\centering
\footnotesize
\renewcommand{\arraystretch}{1.25}
\begin{tabular}{@{}>{\raggedright\arraybackslash}p{2.2cm}p{8.4cm}>{\raggedright\arraybackslash}p{2.5cm}@{}}
\toprule
Wars at the end & Scenario & Happened \\
\midrule
Iran--Israel & Israel masses forces against Iran and strikes it & Iran--Israel: April and October 2024, 13 to 24 June 2025 \\
\hline
Iran--US & Iran masses forces against the United States and strikes it & Iran--US: 22 to 24 June 2025 \\
\hline
Iran--Ukraine & Iran masses forces against Ukraine and strikes it, then masses forces against the United States & no \\
\hline
Iran--Ukraine, Iran--US & Iran masses forces against Ukraine and strikes it; Iran and the United States mass forces against each other; the United States strikes Iran & Iran--US: 22 to 24 June 2025 \\
\hline
Iran--Israel, Iran--US & Iran masses forces against Israel and strikes it; Iran and the United States mass forces against each other; the United States strikes Iran & Iran--Israel: April and October 2024, 13 to 24 June 2025; Iran--US: 22 to 24 June 2025 \\
\hline
Russia--US & Russia imposes targeted sanctions on the United States; the United States appeals to Iran for support against Russia; Russia threatens the United States; the United States masses forces against Russia and strikes it & no \\
\hline
Iran--US, Russia--US & Russia threatens the United States; the United States masses forces against Russia and against Iran, then strikes Iran and strikes Russia & Iran--US: 22 to 24 June 2025 \\
\hline
Russia--Ukraine & Russia presses demands on Ukraine; Ukraine appeals to Iran for support against Russia; Ukraine threatens Russia; Russia masses forces against Ukraine and attacks it & Russia--Ukraine: from 24 February 2022 \\
\bottomrule
\end{tabular}
\caption{Scenarios use a $5$ budget as of 1 January 2021. 1 per reachable outcome of $\driverset$. Each outcome is the set of pairs at war at the end. The last column shows which wars in the outcome occurred between 1 January 2021 and 1 January 2026, based on dated news reports.}
\label{tab:scenarios}
\end{table}

\paragraph{\textbf{The scenario set (Step 4).}}
\Cref{tab:scenarios} lists the possible outcomes.
5 pairs engage in war in at least 1 scenario: Iran with Israel, Ukraine and the United States, and Russia with Ukraine and the United States.
At this budget, no scenario shows China going to war with any country, Russia with Iran or Israel, or Israel, Ukraine and the United States with each other.
Every war in the set needs a mobilisation and a strike, preceded by a threat when the pair was not already hostile.
A larger budget would allow for such outcomes but would require more processing time.
We evaluate each pair based on news coverage, defining a pair as at war when both sides strike the other's forces or territory.
This definition is broader than the board's, which labels a pair in conflict only after 280 uses of force in each direction within 1 week of machine-coded news.
We rely on the record because that week of news cannot be read 5 years later, so a 2-day exchange of strikes counts as a war here but not for the board.
3 pairs went to war within the window.
Russia invaded Ukraine on 24 February 2022, after accumulating forces along its border in the spring and autumn of 2021.
This conflict persisted through the end of the observation period.\footnote{\raggedright \url{https://en.wikipedia.org/wiki/Prelude_to_the_Russian_invasion_of_Ukraine} and \url{https://en.wikipedia.org/wiki/Russian_invasion_of_Ukraine}, accessed 28 September 2026.} Iran and Israel exchanged strikes on each other's territories on 13 to 14 April and 19 April 2024, and again on 1 and 26 October 2024.\footnote{\raggedright \url{https://en.wikipedia.org/wiki/April_2024_Iranian_strikes_against_Israel}, \url{https://en.wikipedia.org/wiki/April_2024_Israeli_strikes_on_Iran}, \url{https://en.wikipedia.org/wiki/October_2024_Iranian_strikes_against_Israel}, \url{https://en.wikipedia.org/wiki/October_2024_Israeli_strikes_on_Iran}, accessed 28 September 2026.} Israel struck Iran again on 13 June 2025, and the United States targeted 3 of Iran's nuclear sites on 22 June.
Iran retaliated by firing missiles at a U.S. base in Qatar on 23 June, with a ceasefire established the following day.\footnote{\raggedright \url{https://en.wikipedia.org/wiki/Twelve-Day_War} and \url{https://en.wikipedia.org/wiki/United_States_strikes_on_Iranian_nuclear_sites}, accessed 28 September 2026.} All 3 pairs are included in the 5 pairs recorded as being at war, with the first, second and eighth rows of \Cref{tab:scenarios} representing scenarios involving each.
The other 2 pairs the set puts at war (Iran with Ukraine, and Russia with the US) did not fight, and neither did the remaining pairs the set leaves at peace.
The 3 wars overlapped in time, and the set holds 1 overlap but not the other 2.
The war between Russia and Ukraine began on 24 February 2022.
This war coexisted with the conflict between Iran and Israel during their exchanges, as well as with the war between Iran and the United States from 22 to 24 June 2025.

\expandafter\def\expandafter\UrlBreaks\expandafter{\UrlBreaks\do\-}
\subsection{Supply Chains}\label{sec:case-study:supply-chain}

A shipping-route disruption delays parts that manufacturers source internationally, affecting container lines, suppliers, and factories.
This analysis focuses on 4 key routes: the Suez Canal, the Panama Canal, the port of Baltimore, and the Strait of Hormuz, and covers incidents from 1 January 2021 to 1 January 2026.
We use action schemas from supply-chain risk literature, rather than an event taxonomy, based on pre-2021 sources.

\paragraph{\textbf{The planning domain (Step 1).}}
The planning domain includes 17 action schemas covering container lines, the \emph{maker} (the manufacturer whose question the board answers), rival buyers, their supplier, and route-keeping authorities, as well as 6 happenings.
The happenings are 5 risk categories from \citet{chopra2004managing} (accident blockage, ship attack, drought/low water, demand surge, late delivery) plus closure by force, following \citet{talmadge2008closing} on Iran and the Strait of Hormuz (a wide strait cannot be closed by a single accident).
Buyer moves follow \citet{tomlin2006value}: use stock, fly parts, shift orders to a second source, or accept the loss and stop a line.
Buyers can inflate orders (the \emph{rationing game} of \citet{lee1997information}); suppliers can ration and announce output.
Lines can reroute, cancel sailings, or add a surcharge, following \citet{notteboom2006time}.
Authorities can clear blockages (recovery in \citet{sheffi2005supply}) or escort ships, as in Operation Atalanta since 2008.\footnote{\raggedright \url{https://en.wikipedia.org/wiki/Operation_Atalanta}, accessed 26 September 2026.} Fluents track whether a route is closed or under attack, whether lines have diverted, and whether a buyer is short of parts, has inflated its order, or has stopped a line.
4 numeric fluents range from 0 to 3: route capacity, lane lateness, freight rates, and buyer stock.
Most actions shift 1 level: blockage or forced closure shuts the route and makes the lane 2 levels late; a late delivery leaves the buyer short; drawing on stock covers the shortage but reduces stock by 1 level.
This cost is coarse-grained, since a late delivery counts as 1 event regardless of duration.

\paragraph{\textbf{The question and topic (Step 2).}}
The analyst asks how a disruption could affect the parts stock of a European manufacturer whose parts ship from Asia via the Suez Canal.
The analyst asks the same for manufacturers supplied via the Panama Canal and the port of Baltimore, and for a refiner whose crude ships through the Strait of Hormuz.
Because the concern is stock, a scenario matters only if the manufacturer's stock ends at a different level than it began. The concern $\concernset$ is the maker's stock.
The drivers are (1) whether the canal is still closed at the end, (2) whether container lines divert, and (3) whether the manufacturer halts a production line.
We chose these because post-event traffic data identify the first 2, and a dated announcement identifies the third. 
Lines want ships moving and rates up; buyers want stock up, no missed deliveries, and lines operating.
Suppliers want only fillable orders; authorities want open, safe routes.
Late deliveries can recur because each missed sailing creates another delay, and buyers may repeatedly draw on stock.

\begin{lstlisting}[caption={Part of the supply-chain topic, which holds no law (the full file is in the supplementary code).},label={lst:supply-topic}]
#program base.
happening(A) :- actArg(A, 0, W), has(W, type("nature")).
repeatable(A) :- actClass(A, "late_arrivals").
sustainedClass("draw_stock").
preference(B, variable(("short",B)), false, 3) :- buyer(B).
preference(B, variable(("inventory",B)), up, 1) :- buyer(B).
preference(S, variable(("short_supply",S)), false, 1) :- supplier(S).
concern(variable(("inventory",B))) :- focal(B).
driver(variable(("line_paused",B))) :- focal(B).
\end{lstlisting}

\paragraph{\textbf{The initial state (Step 3).}}
The script reads 3 sources for each route and date: IMF PortWatch, the Federal Reserve Bank of New York's Global Supply Chain Pressure Index (GSCPI), and the GDELT Global Knowledge Graph.
IMF PortWatch tracks daily ship traffic, and the script deems a route closed when last-day traffic drops below 35\% of baseline (the mean over the 28 days ending 1 week earlier) and GDELT records at least 500 articles on maritime incidents on that route in the past 2 days\footnote{Both thresholds are ours.}.
The GSCPI indicates rising freight rates when it exceeds 1 standard deviation above the mean, while GDELT categorises articles by theme and location.
An attack is indicated if $300$ weekly articles on maritime incidents feature conflict or terror themes and constitute at least $25\%$ of the total.
The model's thresholds are based on the Ever Given grounding in March 2021, when traffic fell to $24\%$ of baseline.
This event falls within the evaluation window, so we report its outcome (\Cref{tab:supply-scenarios}) but do not count it among the outcomes.
In contrast, traffic remained above $62\%$ before the incident, preventing misclassification of regional tensions as attacks.
On 1 January 2021, all routes were operating at nearly full capacity, with no attacks reported and freight rates indicating a stable supply situation.

\paragraph{\textbf{The scenario set (Step 4).}}
\Cref{tab:supply-scenarios} reports Suez-board outcomes: route closed, lines diverted, or both, with the maker's lines running.
No scenario stops a line: scenarios contain at most 5 events, while a stop requires a disruption, 3 late deliveries, and 2 prior draws on the maker's stock of $2$.
A larger budget could reach stops, but would increase processing time.
We validate each driver against dated news and company announcements.
The route is ``closed'' when shut to shipping; lines are ``diverted'' when container lines reroute around it or to other ports; a line is ``stopped'' when a manufacturer shipping parts via the route reports a disruption-related stoppage.
3 outcomes occurred, of which we count 2.
The Ever Given blocked the canal from 23 to 29 March 2021, and some ships rerouted via the Cape of Good Hope, so $\langle\mathit{true},\mathit{true},\mathit{false}\rangle$ held that week\footnote{We report this outcome and do not count it, since the thresholds were calibrated on it.}.
From mid-December 2023, after attacks on ships,\footnote{\raggedright \url{https://en.wikipedia.org/wiki/Houthi_attacks_on_commercial_vessels} and \url{https://www.voanews.com/a/msc-to-divert-shipping-away-from-suez-canal-after-red-sea-attacks/7400938.html}, accessed 27 September 2026.} container lines diverted away from the Red Sea.
PortWatch still shows them on the longer route at the window's end, so $\langle\mathit{false},\mathit{true},\mathit{false}\rangle$ held thereafter.
In January 2024, Tesla, Volvo Cars, Suzuki, and Michelin halted European plants after diversions delayed parts and materials,\footnote{\raggedright \url{https://money.usnews.com/investing/news/articles/2024-01-11/exclusive-tesla-berlin-to-suspend-most-production-for-two-weeks-over-red-sea-supply-gap}, \url{https://www.marketscreener.com/news/latest/Volvo-Cars-pauses-production-in-Belgium-due-to-Red-Sea-situation-45731919/}, \url{https://www.timeslive.co.za/motoring/news/2024-01-15-hungarys-suzuki-plant-to-stop-production-for-a-week-due-to-supply-chain-issues/} and \url{https://www.tyrepress.com/2024/01/job-cuts-shortages-stress-for-michelin-in-spain/}, accessed 27 September 2026.} so $\langle\mathit{false},\mathit{true},\mathit{true}\rangle$ held in late January 2024.
Of the 2 outcomes, the set includes the diversion (row 2) and the Ever Given outcome (row 1), which misses the stop.
Row 2 reaches the diversion because a blockage the authorities later cleared forced ships to divert.
In Baltimore, a ship hit the Key Bridge on 26 March 2024, closing the port and forcing lines to divert to other ports.
A deep-draught channel reopened on 25 April and the federal channel on 10 June,\footnote{\raggedright \url{https://en.wikipedia.org/wiki/Francis_Scott_Key_Bridge_collapse} and \url{https://www.army.mil/article/277129/usace_announces_full_restoration_of_baltimores_fort_mchenry_federal_channel}, accessed 28 September 2026.} so $\langle\mathit{true},\mathit{true},\mathit{false}\rangle$ held in the interim.
The set reflects this in the record's order: the port is blocked, and lines reroute to the Port of Virginia.
It also posits a demand surge, an inflated order by the maker, and a surcharge that no source records.
On the Panama board, drought reduced canal transits from mid-2023.
From January 2024, Maersk moved 1 service's cargo across Panama by rail,\footnote{\raggedright \url{https://pancanal.com/en/maritime-services/advisory-to-shipping/} and \url{https://www.cnbc.com/2024/01/11/panama-canal-drought-moves-maersk-to-start-using-land-bridge-for-cargo.html}, accessed 28 September 2026.} so $\langle\mathit{false},\mathit{true},\mathit{false}\rangle$ applied to 1 line.
The set captures the outcome but neither its cause nor the response.
For instance, it witnessed the diversion through a blockage the authorities later cleared rather than low water, and it sent the lines around the Strait of Magellan, whereas Maersk used rail, which the domain lacks.
On the Hormuz board, none of the 3 outcomes occurred.
Across the 4 boards, 1 outcome occurred at Suez, Panama, and Baltimore. 
The other outcomes did not occur (1 at Suez, 2 at Panama, 2 at Baltimore, and 3 at Hormuz), and the sets miss the stopped lines at Suez.

\begin{table}[t]
\centering
\footnotesize
\begin{tabular}{@{}cccp{7.9cm}>{\raggedright\arraybackslash}p{2.2cm}@{}}
\toprule
\multicolumn{3}{c}{Outcome} & Scenario & Happened \\
closed & diverted & paused & & \\
\midrule
true & true & false & demand for the parts surges; the canal is blocked; the lines reroute round the Cape of Good Hope; the maker inflates its order; the lines add a surcharge & yes: 23--29 Mar 2021, the Ever Given\\ \hline
false & true & false & demand for the parts surges; the canal is blocked; the lines reroute round the Cape; the maker inflates its order; the authorities clear the canal & yes: from mid-Dec 2023 to the end of the window \\ \hline
true & false & false & demand for the parts surges; the rivals inflate their orders; the canal is blocked; the maker inflates its order; the lines add a surcharge & no \\
\bottomrule
\end{tabular}
\caption{Scenarios for the Suez board, listed by reachable $\driverset$ outcomes. The outcome $\langle\mathit{false},\mathit{false},\mathit{false}\rangle$ is excluded. The last column indicates whether the outcome occurred between 1 January 2021 and 1 January 2026, based on news reports and company announcements.}
\label{tab:supply-scenarios}
\end{table}

%% file: sections-v5/04b-related-work.tex
\section{Related Work}\label{sec:related-work}

In \Cref{sec:background:scenario-planning}, we noted that an analyst cannot write out every trajectory of a topic by hand, and \Cref{sec:engine,sec:case-study} presented an engine that does so and applied it to two topics.
In this section, we relate the engine to earlier efforts to automate scenario planning.
These efforts fall into four categories: assembling driver states based on consistency, speeding up the modelling tasks involved in the manual process, producing scenarios using planning technologies, and creating scenario drafts with generative models.

\textbf{Consistency-based combination.}
The earliest strand focuses on automating end-state selection.
\citet{weimerjehle2006cross} present cross-impact balance analysis, which evaluates every combination of driver states using pairwise influences elicited from analysts and retains only combinations with internally consistent scores.
\citet{ritchey2006problem} maps the same combinatorial space through morphological analysis and reduces it using pairwise consistency assessments.
\citet{bradfield2005origins} review cross-impact analysis, from which cross-impact balances descend, and the intuitive-logics approach initiated by Shell, treating them as distinct schools within the same broader practice.
In both approaches, a consistency score specifies what qualifies a particular combination of driver states as a scenario.
Our approach applies consistency constraints to every narrative state, and a combination of driver states counts as an outcome only when a plan reaches it within budget.

\textbf{Accelerating the manual exercise.}
The second group keeps analysts in charge while automating model construction.
\citet{davis2023towards} reviews the methods proposed for automating the construction of simulation scenarios.
\citet{thesis-scenario-acceleration} speeds up scenario development by using natural language processing to extract a fuzzy cognitive map from documents, enabling the simulation of quantitative scenarios based on that map.
These approaches depict the world as a network of interconnected drivers, so a generated scenario illustrates how various trends can reinforce or dampen one another.
Our scenarios are instead sequences of events that actors cause, together with the states these events lead to.
We keep analysts in control of the model (i.e., they provide the actions, constraints, and drivers), and we automate narrative generation, although our finite value ranges can only approximate the gradual changes a cognitive map expresses directly. 

\noindent\textbf{Planning-based generation.}
The closest system to ours is the IBM Scenario Planning Advisor (SPA)~\cite{sohrabi2018aaai,sohrabi2018ibm,feblowitz2021ibm}.
\citet{sohrabi2018aaai} develop a model of causal forces by gathering analyst insights and then compiling them into a planning problem.
In this model, action costs are derived from the elicited likelihoods, so a cheaper plan represents a more likely explanation of the observed risk drivers.
They produce a set of high-quality plans using top-$k$ planning~\cite{katz2018topk,katz2020topquality}, then cluster them into scenarios~\cite{sohrabi2018aaai,feblowitz2021ibm}.
Our approach differs from SPA in three ways. First, while SPA ranks plans by likelihood, we assess plausibility without assigning costs or probabilities.
Second, SPA's plans explain observed risk drivers, whereas ours start from a stated present and operate within a set budget.
Third, SPA groups similar plans after planning, while our criteria are integral to the planning process and guided by the analyst's drivers.
Within a similar domain, \citet{smith2024automated} create training scenarios using narrative planning.
Their goal is skill acquisition rather than exploring alternative futures.
Political-military wargaming produces adjudicated event sequences that more closely resemble our scenarios than driver-based scenarios do~\cite{perla1990art}.
In this context, a human adjudicator determines which moves are permissible, similar to the preconditions and constraints of our planning problem.
We borrow narrative planning's conditions for a believable story, intentional characters, and causally connected events~\cite{riedl2010narrative}, along with Sabre's reading of intentions through character-level utility~\cite {ware2021sabre}.
However, we differ in what we search for, which is narrative planning aims for a specific outcome set by the author, while we explore possible futures without a defined goal.

\noindent\textbf{Generation with large language models.}
The last group drafts scenarios directly with large language models (LLMs).
For instance, \citet{fi17010048} incorporate generative tools into a participatory foresight exercise to create future scenarios, while LLMScenario uses a language model to generate driving scenarios~\cite{llmscenario}.
In our approach, a narrative is an executable plan, and a scenario is a narrative that meets the criteria; neither is something a language model can guarantee, nor does either read as fluent prose, and both require a planning model of the topic.

Among these four categories, only planning-based systems return a sequence of events, and none state a criterion for when the whole sequence counts as a scenario.
The scenario planning problem in \Cref{def:scenario-problem} provides that criterion, and the \Cref{sec:engine} engine solves it in one solver call.

%% file: sections-v5/05-conclusion.tex
\section{Discussion, Conclusion \& Future Work}\label{sec:conclusion}

\paragraph{\textbf{Discussion.}}
In this paper, we frame scenario planning as a diversity planning problem and introduce an engine that generates one scenario for each potential outcome of a topic within a specified budget.
The approach gives the analyst three key benefits.
First, it presents every scenario as a narrative.
This lets the analyst understand how a future unfolds, not just where it ends.
Second, it applies the criteria uniformly as hard constraints across all narratives, with no ranking, so familiar futures do not get an advantage over unfamiliar ones.
Third, the analyst can rerun it whenever conditions change, since a script rebuilds the board from the day's sources, and the same topic then returns a new set.
The budget bounds what any run can return, since each outcome needs a chain of events that fits within it.
A narrative within a given budget also fits within any larger budget, so the set can only expand as the budget increases.
For instance, the conflict set grows from 5 outcomes at budget 4 to 8 at budget 5, \sm{9} at budget 6 and 11 at budget 7 (\Cref{tab:runs}).
The budget-7 set adds 3 outcomes to the 8 in budget 5, each holding two wars that five events could not, including the Russia--Ukraine war and the Iran--US war, which were underway together in June 2025.
Therefore, a small budget misses outcomes that require longer chains, whereas a larger one reaches more of them at the cost of time and memory (\Cref{tab:runs}).

The engine can return only what the topic models:
it searches the narratives the model allows, so an event that is not modelled never appears in a scenario, however often it occurs in reality.
For example, because the supply-chain domain lacks a rail land bridge, the Panama set cannot represent the 2024 diversion.
Constructing a topic is also much easier when a taxonomy exists.
The conflict domain adheres to CAMEO, allowing each of its 65 schemas to represent one event type, while the same taxonomy organises the news that informs its board.
In contrast, the supply-chain domain lacks such a taxonomy, so we assembled its 17 schemas from five different sources and determined their preconditions and effects ourselves.
A topic built this way records the analyst's every choice, and the set inherits those choices.

\paragraph{\textbf{Conclusion.}}
In this paper, we show that scenario planning can be framed as a diversity planning problem, in which two scenarios differ when they end in different outcomes on the analyst's drivers.
Polyscene encodes scenario criteria as trajectory constraints and returns a set containing one scenario for each possible outcome the topic can achieve within a given budget.
In return, the analyst must model the topic accurately, and a small budget narrows the set to the outcomes that short chains can reach.
We applied Polyscene to geopolitical conflicts and shipping-route disruptions as of 1 January 2021.
With a $5$ budget, the conflict set puts five of the $15$ pairs at war.
By 1 January 2026, three of those wars occurred (Russia--Ukraine, Iran--Israel, Iran--U.S.), and each ends a scenario.
None of the ten pairs left at peace at this budget fought, and neither of the other two pairs marked at war did.
On the supply-chain boards, the sets hold the Red Sea diversion, the closure and diversion at Baltimore, and the partial diversion at Panama.

\paragraph{\textbf{Future work.}}
Two directions remain.
First, prune reorderings of independent events to reduce time and memory costs for larger budgets.
Many narratives differ only in the order of independent events, so the solver explores multiple orders that yield the same outcome.
For instance, in the fifth row of \Cref{tab:scenarios}, Iran and the United States mobilise; because the mobilisations read and write different fluents, either can occur first and both yield the same result.
Differentiation keeps one narrative per outcome, so the engine returns only one, but the solver still searches all orders, which grow factorially with the number of independent events. Second, let the analyst choose the narrative representing an outcome.
Let a topic leave some initial unknowns when the record provides no information about them. Currently, the script sets a specific value for each unknown, and each scenario uses that fixed value. In assumption-based planning~\cite{davismendelow2013assumption}, a planner can choose a value for an unknown and include this assumption in the plan. A scenario could then depend on this stated assumption rather than a value the script decided. This approach could lead to a wider range of possible outcomes based on that assumption, but it requires the reader to check both the assumptions and the record for accuracy.